%% file: main.tex
\documentclass[]{gmaart2019}

\renewcommand{\varProcName}{Proc. 31. Workshop Computational Intelligence}
\renewcommand{\varLocation}{Berlin}
\renewcommand{\varDate}{25.-26.11.2021}

\usepackage[utf8]{inputenc} 
\usepackage[T1]{fontenc} 
 
\usepackage{amsmath,amssymb,amsfonts,mathtools,amsthm} 

\usepackage[babel,autostyle]{csquotes}

\usepackage{tabularx}
\newcolumntype{L}[1]{>{\raggedright\arraybackslash}p{#1}} 
\newcolumntype{C}[1]{>{\centering\arraybackslash}p{#1}} 
\newcolumntype{R}[1]{>{\raggedleft\arraybackslash}p{#1}} 
\usepackage{booktabs}

\usepackage{paralist}   

\usepackage[noend]{algpseudocode}

\algrenewcommand\algorithmicrequire{\textbf{Voraussetzung:}}
\algrenewcommand\algorithmicensure{\textbf{Abschlussbedingung:}}

\usepackage{listings} 
\usepackage{subcaption}  

\usepackage{epstopdf}
\usepackage{xcolor}
\selectcolormodel{gray}  

\usepackage{scrhack}
\usepackage{varwidth}
\usepackage{rotating} 
\usepackage[defaultlines=2,all]{nowidow}  
\usepackage{import}
\usepackage{placeins}


\input{custompackages.tex}

\begin{document}


\hyphenpenalty=2000

\setcounter{page}{1}
\pagestyle{scrheadings}
\pagenumbering{arabic}

\setnowidow[2]
\setnoclub[2]


\input{beitrag}


\end{document}

%% file: custompackages.tex
\usepackage[version-1-compatibility]{siunitx} 

\usepackage{moresize}
\usepackage{adjustbox}
\AppendGraphicsExtensions{.tif}

\usetikzlibrary{shapes.geometric}
\usetikzlibrary{fit,matrix}
\tikzset{database/.style={cylinder,aspect=0.5,draw,rotate=90,path picture={
			\draw (path picture bounding box.160) to[out=180,in=180] (path picture bounding
			box.20);
			\draw (path picture bounding box.200) to[out=180,in=180] (path picture bounding
			box.340);
}}}

\tikzset{
	desicion/.style={
		diamond,
		draw,
		text width=3em,
		text badly centered,
		inner sep=0pt
	},
	block/.style={
		rectangle,
		draw,
		text width=10em,
		text centered,
		rounded corners
	},
	cloud/.style={
		draw,
		ellipse,
		minimum height=2em
	},
	descr/.style={
		fill=white,
		inner sep=2.5pt
	},
	connector/.style={
		-latex,
		font=\scriptsize
	},
	rectangle connector/.style={
		connector,
		to path={(\tikztostart) -- ++(#1,0pt) \tikztonodes |- (\tikztotarget) },
		pos=0.5
	},
	rectangle connector/.default=-2cm,
	straight connector/.style={
		connector,
		to path=--(\tikztotarget) \tikztonodes
	}
}

\newlength{\Oldarrayrulewidth}

%% file: beitrag.tex
\renewcommand{\Title}{Label Assistant: \\A Workflow for Assisted Data Annotation in Image Segmentation Tasks}

\renewcommand{\Authors}{Marcel P. Schilling\textsuperscript{1}, Luca Rettenberger\textsuperscript{1}, Friedrich Münke\textsuperscript{1}, \\Haijun Cui\textsuperscript{2}, Anna A. Popova\textsuperscript{2}, Pavel A. Levkin\textsuperscript{2}, \\Ralf Mikut\textsuperscript{1}, Markus Reischl\textsuperscript{1}}
\renewcommand{\Affiliations}{\textsuperscript{1}Institute for Automation and Applied Informatics \\
		\textsuperscript{2}Institute of Biological and Chemical Systems\\
		Karlsruhe Institute of Technology\\
		Hermann-von-Helmholtz-Platz 1, 76344 Eggenstein-Leopoldshafen\\
		E-Mail: marcel.schilling@kit.edu}

							 
\renewcommand{\AuthorsTOC}{M. P. Schilling, L. Rettenberger, F. Münke, H. Cui, A. A. Popova, P. A. Levkin, R. Mikut, M. Reischl} 
\renewcommand{\AffiliationsTOC}{Institute for Automation and Applied Informatics/Institute of Biological and Chemical Systems, Karlsruhe Institute of Technology} 

\setLanguageEnglish
							 
\setupPaper[Label Assistant: \\A Workflow for Assisted Data Annotation in Image Segmentation Tasks] 


\section*{Abstract}
Recent research in the field of computer vision strongly focuses on deep learning architectures to tackle image processing problems. Deep neural networks are often considered in complex image processing scenarios since traditional computer vision approaches are expensive to develop or reach their limits due to complex relations. However, a common criticism is the need for large annotated datasets to determine robust parameters. Annotating images by human experts is time-consuming, burdensome, and expensive. Thus, support is needed to simplify annotation, increase user efficiency, and annotation quality. In this paper, we propose a generic workflow to assist the annotation process and discuss methods on an abstract level. Thereby, we review the possibilities of focusing on promising samples, image pre-processing, pre-labeling, label inspection, or post-processing of annotations. In addition, we present an implementation of the proposal by means of a developed flexible and extendable software prototype nested in hybrid touchscreen/laptop device.

\section{Introduction}
\emergencystretch 3em
Current research in the domain of image processing is focused on Deep Learning (DL) architectures. Deep Neural Networks (DNNs) like for instance Convolutional Neural Networks (CNNs) show very promising results to solve computer vision tasks like image classification or segmentation. For example, AlexNet~\cite{Krizhevsky2012ImageNetClassificationDeep} with more than 80.000 citations (date of statistic: May, 2021) w.r.t. image classification on ImageNet~\cite{Deng2009ImageNetLargescaleHierarchical} shows the impact of DL in the field of image processing. Walsh et al.~\cite{Walsh2019DeepLearningVs} argue that DNNs are beneficial to achieve accurate prediction quality in complex scenarios like biomedical applications. 

However, the authors in \cite{Walsh2019DeepLearningVs, Chi2020DeepLearningbasedMedical} name as one general bottleneck of DL that image annotation\footnote{Label and annotation are used as a synonym in this article.} is time-consuming and often requires expert knowledge as a bottleneck. Besides, following the arguments of Northcutt et al.~\cite{Northcutt2021PervasiveLabelErrors}, label quality can negatively affect model performance. This may lead to a selection of sub-optimal machine learning models since benchmarks with errors in labels are not reliable in general. Karimi et al.~\cite{Karimi2020DeepLearningNoisy} argue that especially in small data scenarios like biomedical problems, an erroneous annotation may significantly reduce the performance of DNNs.

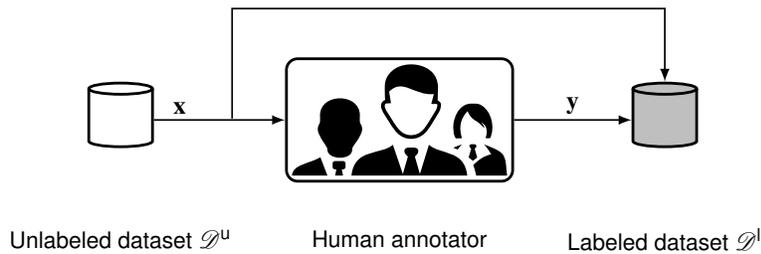
\begin{figure}
	\centering
	\resizebox{0.6\textwidth}{!}{%
		\input{./figs-src/state-of-the-art-labeling.tex}
	}
	\caption{Na\"ive Workflow: A human annotator iterates over an unlabeled dataset $\mathcal{D}^{\text{u}}$ to sequentially label a sample $\mathbf{x}$ in order to generate labels $\mathbf{y}$ to build a labeled dataset $\mathcal{D}^{\text{l}}$ without any form of assistance.}
	\label{fig:naive-workflow}
\end{figure}
The na\"ive way to generate a labeled dataset $\mathcal{D}^{\text{l}} = \lbrace (\mathbf{x}_i, \mathbf{y}_i)  \mid i=1,\ldots,M \rbrace$ composed of $M$ instances is represented in Figure~\ref{fig:naive-workflow}. An annotator adds sequentially corresponding labels $\mathbf{y}_i$ to samples $\mathbf{x}_i$ of the unlabeled dataset $\mathcal{D}^{\text{u}} = \lbrace \mathbf{x}_i \mid i=1,\ldots,N \rbrace$ assembled of $N\geq M$ instances without any form of assistance. The labeled dataset $\mathcal{D}^{\text{l}} $ incrementally increases during labeling.

There are several ideas to enhance image annotation for the development of DL applications w.r.t. decreasing annotation effort and improving annotation quality which will be presented as an overview in Section~\ref{sec:sota}. 

Current research predominantly focuses on separate aspects of ways to enhance a na\"ive generation of annotated datasets. However, to the best of our knowledge, there is no generic workflow summarizing and combing ideas of improving the image annotation procedure. We are structuring the ideas and thereby propose a comprehensive workflow. The proposal is intended to serve as a template that can be used as an initial starting point for DL projects in cases where a labeled dataset for supervised learning is required. 

Our key \textbf{contributions} are the following:
\begin{itemize}
	\item a survey of methods/approaches to assist data annotation for DL,
	\item a generic workflow build on meaningful combinations as well as extensions of them, and
	\item the introduction of a developed and extendable software prototype which can be used for assisted labeling in practical problems.
\end{itemize}
Related work is summarized in Section~\ref{sec:sota}. Our workflow and methods are presented in Section~\ref{sec:method}. Besides, the software implementation is described in Section~\ref{sec:implementation} following obtained results in Section~\ref{sec:results}. Finally, we conclude our work in Section~\ref{sec:conclusion}.
\section{State of the Art}
\label{sec:sota}
The requirement of annotated data is an often addressed issue in the context of supervised DL approaches. Data efficient architectures~\cite{Scherr2020CellSegmentationTracking, Isensee2021NnUNetSelfconfiguringMethod}, self-supervised learning ~\cite{Chen2020SimpleFrameworkContrastive}, semi-supervised learning~\cite{Chen2021SemisupervisedSemanticSegmentation}, and transfer learning~\cite{Tan2018SurveyDeepTransfer} are methods to deal with hurdle of obtaining labeled data from the perspective of network architecture/training. Considering data annotation, there are two aspects to take into account - \textit{labeling effort}~\cite{Walsh2019DeepLearningVs, Chi2020DeepLearningbasedMedical} and \textit{label quality}~\cite{Karimi2020DeepLearningNoisy, Northcutt2021PervasiveLabelErrors}. In general, decreasing manual effort for users while maintaining high label quality is desired.

There are basic \textbf{software packages} like LabelMe~\cite{Wada2016LabelmeImagePolygonal}, Pixel Annotation Tool~\cite{Breheret2017PixelAnnotationTool}, \mbox{Image Labeling Tool}~\cite{Bartschat2019ImageLabelingTool} or the basic release of Fiji/ImageJ~\cite{Schindelin2012FijiOpensourcePlatform} for annotating images in the context of segmentation like depicted as na\"ive workflow in Figure~\ref{fig:naive-workflow}. 

In the context of labeling, \textbf{Deep Active Learning} (DAL) surveyed in~\cite{Ren2020SurveyDeepActive} is proposed as a method to reduce labeling effort. The key concept of the mostly considered pool-based sampling is using a more elaborate sampling strategy in contrast to do a straightforward sequential approach. Based on a criterion, also named as query strategy, the human annotator should focus on the most promising samples instead of annotating without any sampling strategy na\"ively. As depicted in~\cite{Ren2020SurveyDeepActive}, criteria can be in terms of model uncertainty or diversity of the dataset  (e.g. measured via distances in latent feature space). However, DAL research mainly focuses on a theoretical perspective. Implementations in open-source labeling tools like ~\cite{Bartschat2019ImageLabelingTool,Falk2019UnetDeepLearning,Hollandi2020AnnotatorJImageJPlugin,Sekachev2020ComputerVisionAnnotation} lack, only few commercial supplier like Labelbox~\cite{Sharma2021Labelbox} provide interfaces to affect sampling.

A few software tools already have implemented the idea of \textbf{pre-labeling}. The general idea of pre-labeling is using a heuristic as an initial guess to simplify labeling. For instance, the Computer Vision Annotation Tool~\cite{Sekachev2020ComputerVisionAnnotation} or Fiji/ImageJ plugins presented in~\cite{Falk2019UnetDeepLearning,Hollandi2020AnnotatorJImageJPlugin} implement an interface for using deep learning models in order to do image pre-labeling. However, Fiji/ImageJ is implemented in Java and consequently a deployment of models nested in state-of-the-art python-based frameworks like PyTorch~\cite{Paszke2019PyTorchImperativeStyle} or TensorFlow~\cite{Abadi2015TensorFlowLargescaleMachine} requires additional effort. Commercial tools like Labelbox~\cite{Sharma2021Labelbox} also offer an interface to upload pre-labels. Besides, there is a function in terms of automatically creating clusters of pixels based on regional image properties in order to simplify labeling. 
The tool ilastik~\cite{Berg2019IlastikInteractiveMachine} enables semi-automatic image segmentation by a combination of edge detection and watershed algorithm~\cite{Beucher1979UseWatershedsContour}. The authors in~\cite{Englbrecht2021AutomaticImageAnnotation} propose a pipeline for obtaining initial labels based on traditional image processing approaches like Otsu thresholding~\cite{Otsu1979ThresholdSelectionMethod} and watershed algorithm~\cite{Beucher1979UseWatershedsContour}, but an open-source software implementation lacks. Moreover, the tool LabelMe~\cite{Wada2016LabelmeImagePolygonal} offers functionality to use previous neighboring labels as pre-labels which may be beneficial for 3D/spatial or temporal data. 

Furthermore, image \textbf{pre-processing} is another form of assistance in the context of image annotation. For instance, Fiji/ImageJ offers a raw image pre-processing with operations like adjustment of the contrast or noise filtering. The software BeadNet~\cite{Scherr2020BeadNetDeepLearningbased} is an example for image preparation in the sense that images are resampled in order to simplify labeling.

Karimi et al.~\cite{Karimi2020DeepLearningNoisy} and Northcutt et al.~\cite{Northcutt2021ConfidentLearningEstimating} address the issue of noisy labels and survey options to handle them. For instance, the authors in~\cite{Karimi2020DeepLearningNoisy} present methods like pruning wrong labels, adapting DNN structures, developing more elaborate objectives, or changing training procedures to cope with noisy labels. Northcutt et al.~\cite{Northcutt2021ConfidentLearningEstimating} propose Confident Learning, which is a method for pruning wrong labels in a labeled dataset after labeling has finished. Hereby, each sample is ranked concerning the disagreement between predictions of a trained model and corresponding noisy labels. However, the ideas are detached from the actual labeling process and focus on classification. 

In particular, the idea of giving direct \textbf{feedback} concerning segmentation labels is a concept that is not considered in state-of-the-art approaches. Hence, software tools do neither support the possibility of scoring labels w.r.t. quality nor allow post-processing of them. Only some tools like Labelbox~\cite{Sharma2021Labelbox} enable manual tagging of images for a review process in order to allow further manual inspection by other annotators. 

The toolbox LabelMe~\cite{Wada2016LabelmeImagePolygonal} allows using watershed algorithm~\cite{Beucher1979UseWatershedsContour} in order to do \textbf{post-processing} of coarse annotations. However, state-of-the-art tools lack w.r.t. post-processing functions allowing customization depending on the problem.

Moreover, the general approach is that labeling is performed using a mouse as \textbf{input device}. The work of~\cite{Forlines2007DirecttouchVsMouse} compares mouse devices with touch devices. The experiments of the authors show that in case of bimanual tasks, like fitting a mask on an object, touchscreens are beneficial.

The main open problems/questions of related work can be summarized in: (i) no definition of a comprehensive workflow combining different approaches of improving image annotation, (ii) lack of smart methods concerning sample selection directly integrated into the annotation process, (iii) no possibility for direct feedback w.r.t. label quality in the annotation process, and (iv) a missing flexible software implementation to make use of combinations of label assistance.

\section{Methods}
\label{sec:method}
\subsection{Properties and challenges in datasets}
\label{sec:dataset-properties}
In order to introduce a workflow, we give a brief overview of properties in datasets and arising challenges as one part of our contribution:
\begin{itemize}
	\item A dataset may have temporal or spatial relations like videos or 3D images. In this case, neighboring frames are often very similar.
	\item Related to this, datasets composed of video sequences are often very homogeneous within a scene, but quite heterogeneous when comparing different sequences. 
	\item Dealing with for instance microscopy images, areas of interest may be depending on relative changes in gray value/color channels. Thus, not the whole value range in high-resolution images is relevant.
	\item Furthermore, noise in datasets may impede image annotation.
	\item The level of difficulty to solve the task can range from already available heuristics to solve the problem coarsely to hard problems. Here, there are no ways to tackle the problem directly. Besides, within a dataset, there may be a variance in examples w.r.t. difficulty to interpret them.
	\item Depending on the problem, there is often prior knowledge before starting labeling, e.g. a specific number of segments per sample or the desired property of no holes within a segment.
	\item Annotations by humans are not guaranteed to be perfect. Intra-observer and inter-observer variance may lead to errors.
\end{itemize}
The aforementioned properties serve as motivation for following presented approaches and methods included in the workflow proposal (Section~\ref{sec:method-workflow}).
\subsection{Workflow}
\label{sec:method-workflow}
\begin{figure}
	\centering
	\resizebox{\textwidth}{!}{%
		\input{./figs-src/workflow.tex}
	}
	\caption{Assisted Labeling Workflow: A selector chooses promising samples $\mathbf{x}^{*} $ out of the unlabeled dataset $\mathcal{D}^{\text{u}}$. The pre-assistance and post-assistance module guide the human annotator during the labeling procedure. Final labels $\mathbf{y}$ are obtained and the labeled dataset $\mathcal{D}^{\text{l}}$ increases gradually.}%
	\label{fig:workflow}%
\end{figure}
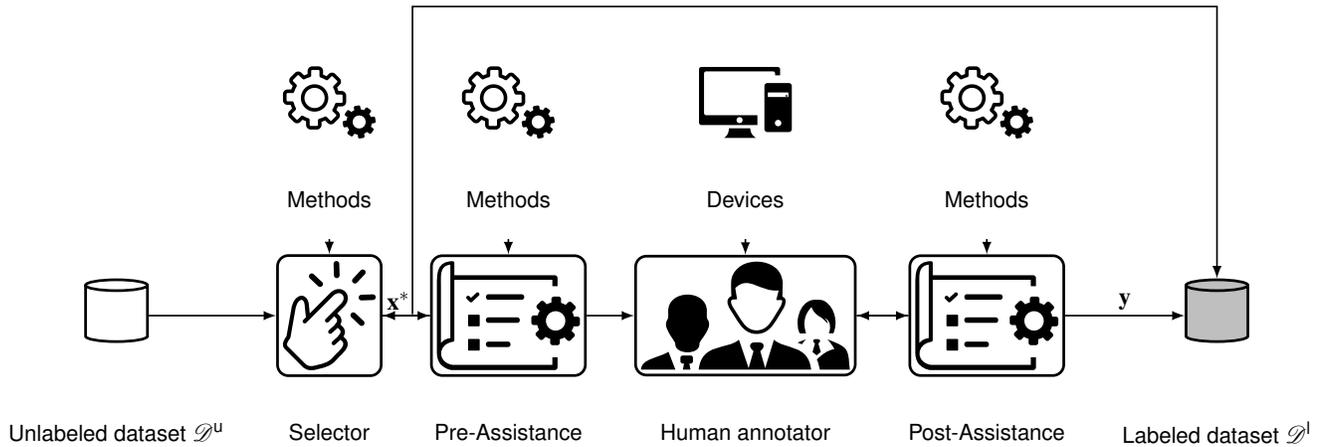
Our proposed workflow is represented in Figure~\ref{fig:workflow}. Firstly, starting from a unlabeled dataset $\mathcal{D}^{\text{u}}$, a selector (cf. Sec.~\ref{sec:method-selector}) prioritizes between all unlabeled samples and favors the next sample to label, denoted as $\mathbf{x}^{*} $. The subsequent pre-assistance module can yield assistance in two ways: providing pre-labels (cf. Sec.~\ref{sec:method-pre-label}) as initial guesses as well as pre-processing of samples (cf. Sec.~\ref{sec:method-pre-pro}) to simplify annotation. Afterward, the labeling is done by the human annotator. This process can be performed using different input devices as depicted in Section~\ref{sec:method-annotator}. Finishing the labeling of the sample, post-assistance is a further part of the workflow. On the one hand, labels can be inspected based on defined metrics in order to provide feedback to the human annotator (cf. Sec.~\ref{sec:method-post-insp}). On the other hand, based on post-processing functions, corrections of the labels are possible (cf. Sec.~\ref{sec:method-post-pro}). Hence, the final label $\mathbf{y}$ is obtained and the number of labeled images in $\mathcal{D}^{\text{l}}$ increases. It should be noted that Figure~\ref{fig:workflow} represents the workflow in total, but in practical applications, the assistance is related to the dataset/task. Hence, in general, not all modules need to be activated.

The following sections are composed of two parts: an introduction of the \textit{concept in general} and \textit{presented methods}. Results of the presented methods can be found in Section~\ref{sec:results}.

\subsection{Selector}
\label{sec:method-selector}
\paragraph{General Concept}
The basic idea of the selector is to allow the user to affect the sampling of images during the labeling procedure and focus on promising samples $\mathbf{x}^{*}$ instead of labeling all images. Let an abstract query strategy, denoted as $a_j \in \mathcal{A}$, be part of the set $\mathcal{A}$ of $A$ query strategies. Thus, $a_j$ takes all unlabeled samples of $\mathcal{D}^{\text{u}}$ into account and maps to a score $s_j(\mathbf{x}) \in \left[ 0,1 \right]$ regarding each sample $\mathbf{x}$. An increasing $s_j(\mathbf{x})$ describes more relevance of a sample. To provide a generic sampling approach, the final score is obtained using weighted averaging 
\begin{equation}
s(\mathbf{x}) = \frac{1}{\sum_{j=1} ^ A w_j^a } \sum_{j=1} ^ A w_j^a
~s_j(\mathbf{x}) 
\label{eq:select}
\end{equation}
based on weights $w_j^a \geq 0$ in order to favor query strategies. The weights $w_j^a \geq 0$ are hyperparameters that need to be obtained depending on the underlying problem and query strategies $a_j$. The next promising sample is obtained by $\mathbf{x}^{*} = \underset{\mathbf{x} \in (\mathcal{D}^u \setminus \mathcal{D}^l)}{\operatorname{argmax}}~s(\mathbf{x})$. 

\paragraph{Presented Methods}
Examples for query strategies in the context of DAL can be found in~\cite{Ren2020SurveyDeepActive}, like for instance using model uncertainty or heterogeneity for sampling. Firstly, we present a novel cherry-picking function for users. The annotator could inspect the dataset and assign  $s_i(\mathbf{x})=1$ for relevant samples $\mathbf{x}$ or $s_i(\mathbf{x})=0$ for images which should not be considered directly at the beginning of the labeling. This clears the hurdle of manually creating a list in parallel, to mark relevant samples. 

Furthermore, we investigate the potential of an automated selector in the context of a sequential dataset. Thereby, we introduce two additional query strategies apart from the traditional ordered sequential sampling. On the one hand, random sampling can serve as a query strategy. On the other hand, we propose a sequence-aware sampler. If the Euclidean difference in reduced gray-level feature space between two images is larger than a pre-defined threshold, a new sequence or strong change within a sequence is detected. Afterward, the sampler selects randomly a sample per cluster and only if each cluster is represented in $\mathcal{D}^{\text{l}}$, a cluster is considered multiple times. It should be remarked, that for complex problems a more elaborate feature reduction method is advantageous.
\subsection{Pre-Assistance}
\subsubsection{Image Pre-processing}
\label{sec:method-pre-pro}
\paragraph{General Concept}
The key idea of image pre-processing is not directly displaying the initial raw image during image annotation. Instead of this, a pre-processed image is generated. Abstractly speaking, the image pre-processing module is a generic function $h$ which yields a pre-processed form of the raw sample $\mathbf{x}$ in terms of
\begin{equation}
\tilde{\mathbf{x}} = h(\mathbf{x}).
\end{equation}
The objective is to accelerate annotation via displaying $\tilde{\mathbf{x}}$ where image understanding is simplified. However, it should always be considered the same pre-processing during labeling a specific dataset since varying image modalities may lead to inconsistent annotation results. 

\paragraph{Presented Methods}
The desired methods are highly correlated to the depicted dataset. Therefore, we limit our presented pre-processing to two example functions $h$: noise filtering to deal with noisy samples and image normalization to handle high-resolution images with relative changes as depicted in Section~\ref{sec:dataset-properties}. Custom functions can be easily implemented to find a solution that is suitable for the individual problem. 

\subsubsection{Pre-labeling}
\label{sec:method-pre-label}
\paragraph{General Concept}
The main idea in the pre-labeling module is utilizing prior knowledge/heuristics, which can serve as an initial guess. Since a correction of labels is in many cases easier than starting labeling from scratch, we propose pre-labeling to boost the annotation of images. Generally speaking, an initial guess 
\begin{equation}
\hat{\mathbf{y}} = l(\mathbf{x})
\end{equation}
is proposed applying a pre-label function $l$. However, it must be considered that pre-labeling is only meaningful if a function exists that solves the problem coarsely. In cases where $l$ predicts mostly wrong labels, correction can slow down annotation in contrast to boost it. To evaluate quality and suitability of a pre-label function, e.g. Dice-Sørensen coefficent~\cite{Jadon2020SurveyLossFunctions}
\begin{equation}
DSC\big(\mathbf{y}(\mathbf{x}),\hat{\mathbf{y}}(\mathbf{x})\big) = \frac{2 \mid \mathbf{y}(\mathbf{x}) \cap  \hat{\mathbf{y}}(\mathbf{x}) \mid }{\mid \mathbf{y}(\mathbf{x}) \mid + \mid \hat{\mathbf{y}}(\mathbf{x}) \mid }
\label{eq:dsc}
\end{equation}
can be utilized as metric comparing initial guess $\hat{\mathbf{y}}(\mathbf{x})$ and ground truth $\mathbf{y}(\mathbf{x})$. Hence, the most suitable pre-label function $l$ or a failure of pre-labeling in total can be determined via \eqref{eq:dsc} evaluating a small set of labeled images.
\paragraph{Presented Methods}
Pre-labeling functions may be various as presented in Section~\ref{sec:sota}. We present several approaches in our software prototype, which can be extended. Firstly, the traditional Otsu segmentation algorithm~\cite{Otsu1979ThresholdSelectionMethod} is shown in order to assist in easier segmentation problems like enumerated in Section~\ref{sec:dataset-properties}. Moreover, we present pre-labeling via DNNs which have already been trained on a subset of labeled samples or datasets of adjacent domains. This is beneficial in difficult image processing problems, where no suitable other heuristic exists. Besides, for sequential datasets (e.g. time-series or spatial relations) a pre-labeling is shown where previous adjacent labels are presented. Though, in this case, only a sequential image sampler is meaningful.  
\subsection{Human Annotator}
\label{sec:method-annotator}
\paragraph{General Concept}
Following the results of Forlines et al.~\cite{Forlines2007DirecttouchVsMouse}, the general idea of the proposed workflow w.r.t. human annotation increases flexibility. Hence, the input device is seen as a selectable parameter of the workflow.
\paragraph{Presented Methods}
The status quo in the context of image annotation is using a mouse as an input device. We present an extension of utilizing a touchscreen for image annotation. Thereby, the touchscreen can be used with a touch pencil and fingers as well to provide a maximum level of flexibility and adaption to annotators' preferences. 
\subsection{Post-Assistance}
\subsubsection{Label inspection}
\label{sec:method-post-insp}
\paragraph{General Concept}
As motivated in Section~\ref{sec:sota}, label inspection addresses noisy labels in datasets. The general idea is to score the annotations based on $G$ metrics $g_j \in \mathcal{G}$ which form a set $\mathcal{G}$. Each metric $g_j$ maps labels $\mathbf{y}$ to quality scores $\gamma_j \in \left[0,1 \right]$. A warning is thrown, if the final weighted score 
\begin{equation}
\gamma (\mathbf{y}) = \frac{1}{\sum_{j=1} ^ G w_j^g } \sum_{j=1} ^ G w_j^g
~\gamma_j(\mathbf{y}) 
\label{eq:gamma}
\end{equation}
falls below a user defined warning threshold $\gamma_0 \in [0,1]$. Analogously to equation \eqref{eq:select}, weights $ w_j^g \geq 0$ allow to prioritize metrics in the final scoring. The user can reinspect the labels in case of $\gamma (\mathbf{y}) \leq \gamma_0$  and errors may be recognized immediately. 
\paragraph{Presented Methods}
Metrics to inspect labels of human annotators can be various. We present in our software prototype methods which rely on expert knowledge. Thereby, we use these priors in combination with region proposals. Thus, the number of holes within a segment or number of segments serve as a quality measure. Thereby, we compare the deviation to a target property defined by an expert (e.g. only one segment per sample). Since the metric is highly correlated to the problem, custom metrics can be implemented to extend the software functionality. Moreover, using predictions of a DNN trained on a small set of labeled data for benchmarking purpose may serve as an alternative approach, which is more generic. However, this is currently not implemented in the prototype. 
\subsubsection{Post-processing}
\label{sec:method-post-pro}
\paragraph{General Concept}
Practical experiments show that some specific errors are reoccurring. In these cases, post-processing can be meaningful. In general, we propose the opportunity to have an abstract post-processing function in the labeling process in order to tackle the problem of noisy labels. Hence, annotators can use this idea in cases where post-processing of labels may be helpful. Displaying a comparison of labels before and after post-processing ensures that assistance is still supervised by human annotators avoiding unwanted changes in post-processing. 
\paragraph{Presented Methods}
We recognized that especially holes or small noisy segments may come up as reoccurring errors. Thus, we implemented morphological operators as a possibility to post-process segmentation maps. Analogously, the post-processing is depending on the dataset and extensions (considering properties like aspect ratio, size, or area) are possible.  
\section{Implementation}
\label{sec:implementation}
The whole generic workflow depicted in Figure~\ref{fig:workflow} is transferred to practical application. Therefore, a software prototype is developed following the modular architecture of the presented workflow in Section~\ref{sec:method}. The proposed concept is implemented in a python package and therefore setup respectively integration via pip is easy to manage for users. Besides, the Graphic User Interfaces (GUIs) are developed using Qt5~\cite{Vestbo2021Qt} and thus are flexible for extensions in order to do further development. We refer to the Image Labeling Tool~\cite{Bartschat2019ImageLabelingTool} for drawing image segmentation masks, since it allows a very flexible way of including pre-labeling without modifying the source code of the tool. Moreover, the publishers provide the tool across different platforms (Linux, Windows). All modules of our proposed workflow include examples concerning processing, scoring, and query functions according to Section~\ref{sec:method}. However, as mentioned, each module allows the implementation of custom functions in order to gain more flexibility. Consequently, users can customize the proposed workflow to the needs being faced with their individual problem respectively dataset. This may boost the application of the workflow prototype in the research community. Especially, the underlying implementation clears the hurdle to connect the proposed workflow with implementations based on state-of-the-art DL frameworks like TensorFlow and PyTorch~\cite{Paszke2019PyTorchImperativeStyle, Abadi2015TensorFlowLargescaleMachine}. 

\begin{figure}
	\centering
	\begin{subfigure}[c]{0.4\textwidth}
		\resizebox{\textwidth}{!}{%
			\includegraphics{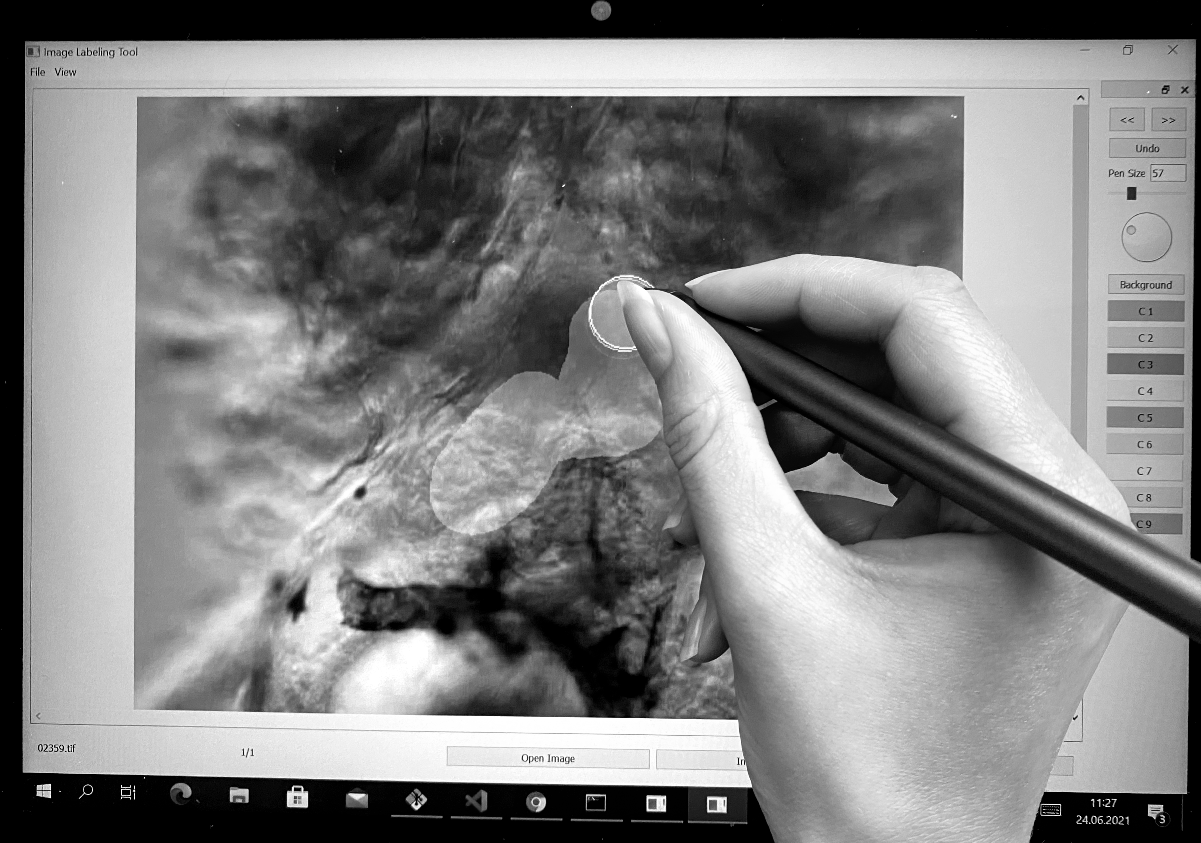}
		}
		\subcaption{Tablet mode with pencil.}
	\end{subfigure}	
	\hfil
	\begin{subfigure}[c]{0.4\textwidth}
		\resizebox{\textwidth}{!}{%
			\rotatebox{-90}{\includegraphics[width=2.55cm, height=3.5cm]{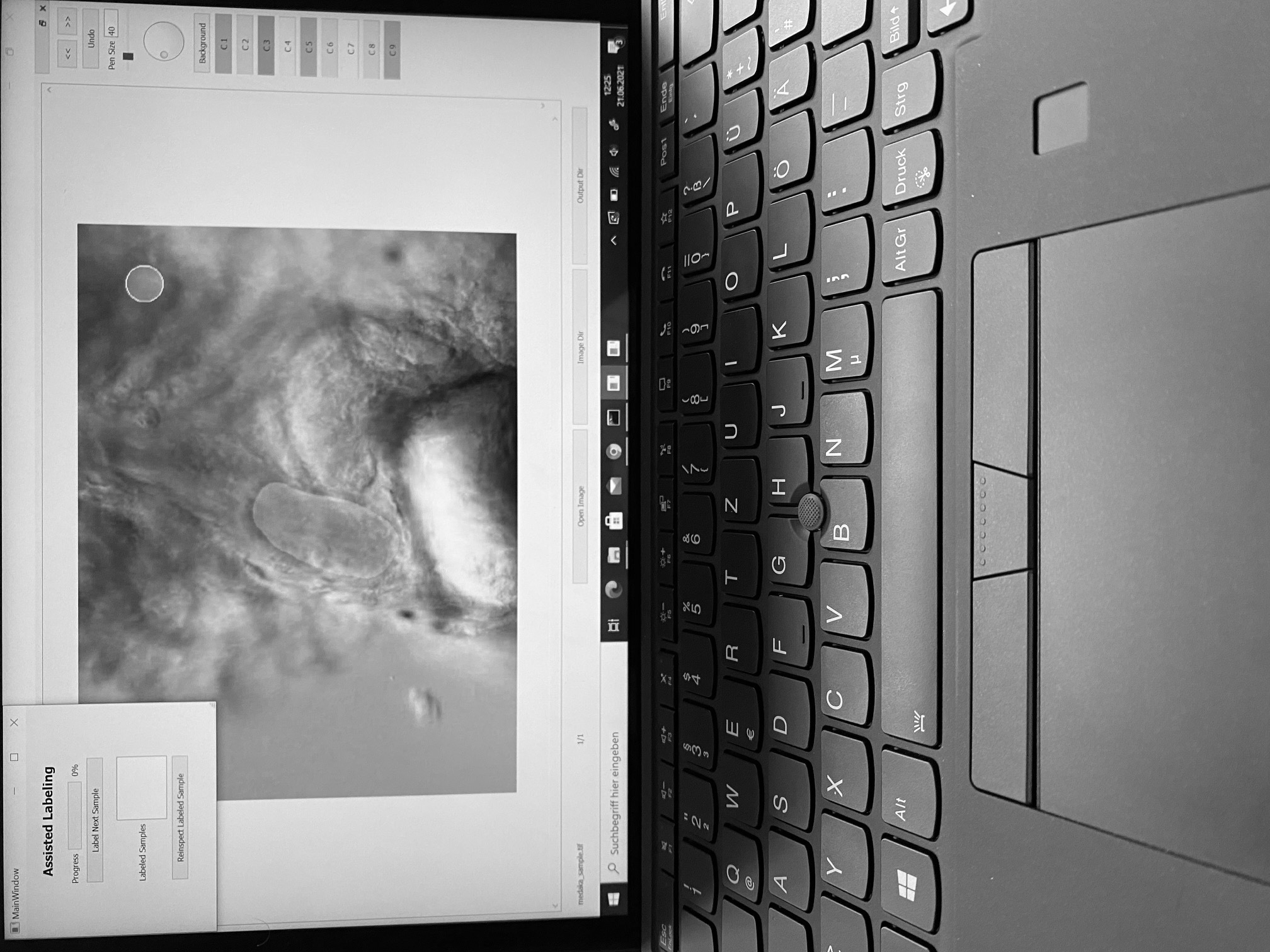}}
		}
		\subcaption{Laptop mode.}
	\end{subfigure}	
	\caption{Software prototype on Lenovo X12 Detachable.}
	\label{fig:tablet}%
\end{figure}

Our software prototype can be used in combination with Windows and Linux operation systems since the implementation is python-based and, using the Image Labeling Tool, relies on a cross-platform segmentation mask drawing tool. We tested it on Windows 10 and Ubuntu 20.04. The system can be used with desktop computers with mouse input devices and tablets as well. Our objective is to provide annotators (e.g. biologists) capsuled hardware, which allows labeling without any installation. Consequently, we deployed our software prototype on a Lenovo X12 Detachable which can be easily handed over to experts as capsuled system. This hardware allows a very flexible usage in terms of offering touch via fingers, touch via a pencil, and laptop mode via keyboard/mouse in parallel. Figure~\ref{fig:tablet} shows the hardware in a practical use-case. 

\section{Results}
\label{sec:results}
\subsection{Datasets}
\label{sec:datasets}
\begin{figure}
	\centering
	\begin{subfigure}[c]{0.495\textwidth}
		\resizebox{\textwidth}{!}{%
			\input{./figs-src/medaka-dataset.tex}
		}
		\subcaption{Medaka}
		\label{fig:medaka}
	\end{subfigure}	
	\hfill
	\begin{subfigure}[c]{0.495\textwidth}
		\resizebox{\textwidth}{!}{%
			\input{./figs-src/dma-dataset.tex}
		}
		\subcaption{DMA Spheroid }
		\label{fig:dma}
	\end{subfigure}	
	\caption{Datasets visualizing exemplary samples and corresponding label masks.}
	\label{fig:data sets}%
\end{figure}
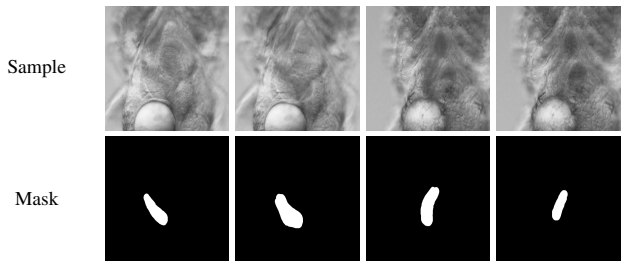
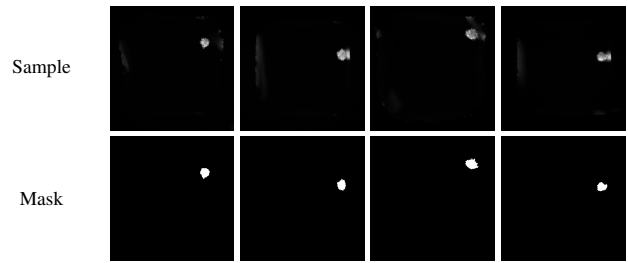
We demonstrate an excerpt of the concept functionalities using two biomedical binary image segmentation datasets depicted in Figure~\ref{fig:data sets}. 
\paragraph{Medaka Dataset} 
The medaka dataset is presented in~\cite{Schutera2019MachineLearningMethods}. It has been released to quantify ventricular dimensions which can be relevant for the understanding of human cardiovascular diseases. An accurate image segmentation of the medaka heart is needed in order to solve this quantification task. The dataset contains 8-bit RGB images and corresponding segmentation masks describing pixels belonging to the ventricle. It includes 565 frames of training data and 165 test samples. Figure~\ref{fig:medaka} illustrates examples and binary segmentation masks. The authors in~\cite{Schutera2019MachineLearningMethods} use the DNN U-Net~\cite{Ronneberger2015UnetConvolutionalNetworks} to solve the image segmentation task. Looking at the example frames, it becomes clear that image segmentation is difficult in this project and thus a simple thresholding algorithm would fail. Furthermore, the dataset is based on roughly 30 video sequences and as presented in Figure~\ref{fig:medaka} neighboring frames may be similar. 
\paragraph{Droplet Microarray Spheroid Dataset} 
The spheroid dataset is recorded in a high-throughput Droplet Microarray (DMA) experiment~\cite{Popova2019FacileOneStep}. Currently, the dataset is not publicly available, a description of the experiment is presented in the work of~Popova et al.~\cite{Popova2019FacileOneStep}. DMA experiments intend to do investigations for drug development and therefore accurate segmentation of fluorescence images is needed. It contains 16-bit high-resolution mono images with corresponding labels obtained by an expert. Thereby, it includes 470 frames of training data and 118 test samples. Being faced with this dataset, the main challenge is to distinguish between artifacts at image boundaries and spheroids. Thus, a straightforward thresholding approach like Otsu~\cite{Otsu1979ThresholdSelectionMethod} is not accurate enough. Figure~\ref{fig:dma} illustrates this problem using example frames respectively segmentation masks. 
\subsection{Experiments}
\paragraph{Selector}
To present the potential of the selector module, we first utilize the medaka dataset introduced in Section~\ref{sec:datasets}, which is a composition of different sequences. In order to evaluate the experiment, we compare $DSC$~\eqref{eq:dsc} using DNN U-Net 
~\cite{Ronneberger2015UnetConvolutionalNetworks} trained on different sampled training datasets (subsets of the initial training dataset) evaluated on a fixed test dataset. The baseline experiment uses almost the entire dataset (400 samples). Hereby, we compare the methods presented in Section~\ref{sec:method-selector}. Results are shown in Table~\ref{tab:selector}.
\begin{table}[t]
	\centering
	\caption{Comparison $DSC$ \iffalse (with standard deviation $\sigma$) \fi of different sampling scenarios (sequential/neighboring, random, sequence-aware) and dataset amounts $\mid\mathcal{D}^l_{\text{train}}\mid$ on medaka dataset~\cite{Schutera2019MachineLearningMethods}.}
	\label{tab:selector}
	\resizebox{0.7\textwidth}{!}{
		\begin{tabular}{ccccc}
			\toprule
			& \multicolumn{4}{c}{\textbf{Configurations}} \\
			& Sequential/neighboring & Random & Sequence-aware & Baseline   \\
			\midrule
			$\mid\mathcal{D}^l_{\text{train}}\mid $& 32 & 32 & 32 & 400   \\
			$DSC$  in \% & 46.50 &77.67&80.63 &82.70 \\
			\bottomrule
		\end{tabular}
	}
\end{table}
A comparison of DNN performance in terms of $DSC$ shows that by considering only a small subset, random sampling and sequence-aware sampling (selecting one random image of each sequence) are superior to standard labeling of neighboring frames in an ordered sequential fashion. However, in this example, the more elaborate sequence-aware approach did not outperform random sampling. If there are no strong imbalances w.r.t. the distribution of the dataset as well as no priors concerning the dataset, random sampling is definitely a proper starting point. Moreover, it can be recognized that the gap from an amount of \mbox{$\mid\mathcal{D}^l_{\text{train}}\mid = 32$}  training samples to the baseline with 400 samples is comparatively small. Hence, with an adapted sampling strategy a small amount is sufficient to obtain accurate results shown by a $DSC > 80 \%$.
\begin{figure}[t]
	\centering
	\begin{subfigure}[c]{0.2\textwidth}
		\resizebox{\textwidth}{!}{%
			\adjustbox{trim={.4\width} {.3\height} {0.1\width} {.2\height},clip}%
			{\includegraphics[width=1.5cm,height=1.5cm]{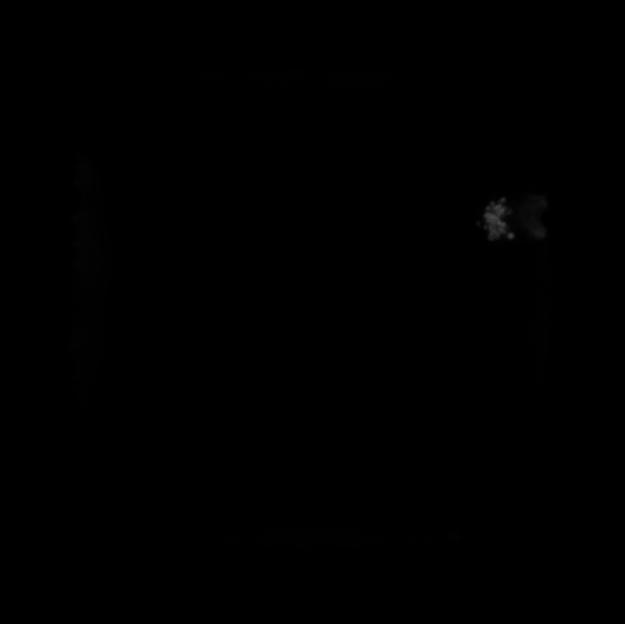}}
		}
		\subcaption{Raw.}
	\end{subfigure}	
	\begin{subfigure}[c]{0.2\textwidth}
		\resizebox{\textwidth}{!}{%
			\adjustbox{trim={.4\width} {.3\height} {0.1\width} {.2\height},clip}%
			{\includegraphics[width=1.5cm,height=1.5cm]{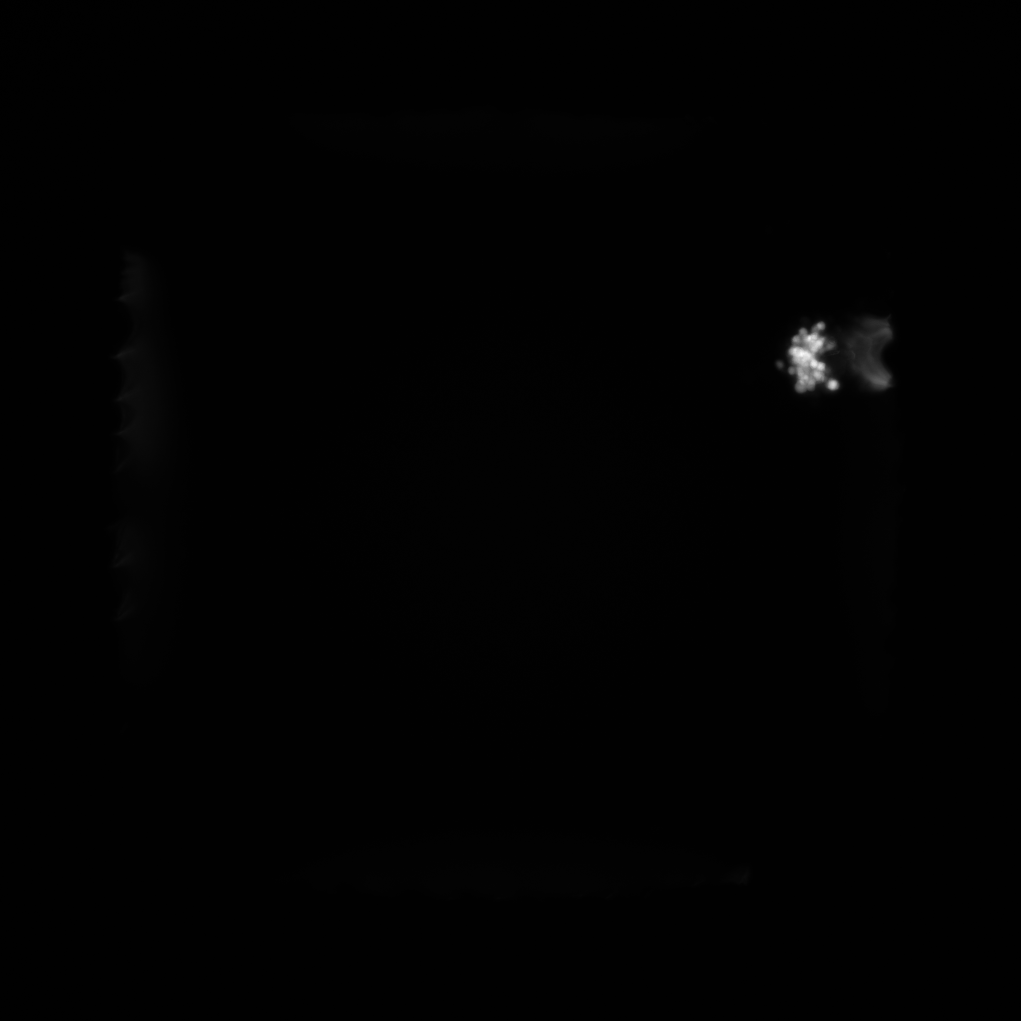}}
		}
		\subcaption{Pre-processed.}
	\end{subfigure}	
	\caption{Example pre-processing on DMA data: a raw sample (a) is processed to a normalized image (b) to enhance image understanding.}
	\label{fig:pre-assist}%
\end{figure}
\paragraph{Pre-processing}
To get an impression of pre-processing, Figure~\ref{fig:pre-assist} represents an example of the DMA spheroid dataset. Thereby, a raw high-resolution DMA mono image is compared to a pre-processed sample. The pre-processing function normalized the gray levels in the image. Thus, relative changes are visible, image understanding is enhanced, and therefore annotating segmentation masks is simplified. 
\paragraph{Pre-labeling}
\begin{figure}[t]
	\centering
	\resizebox{.85\textwidth}{!}{%
		\includegraphics[]{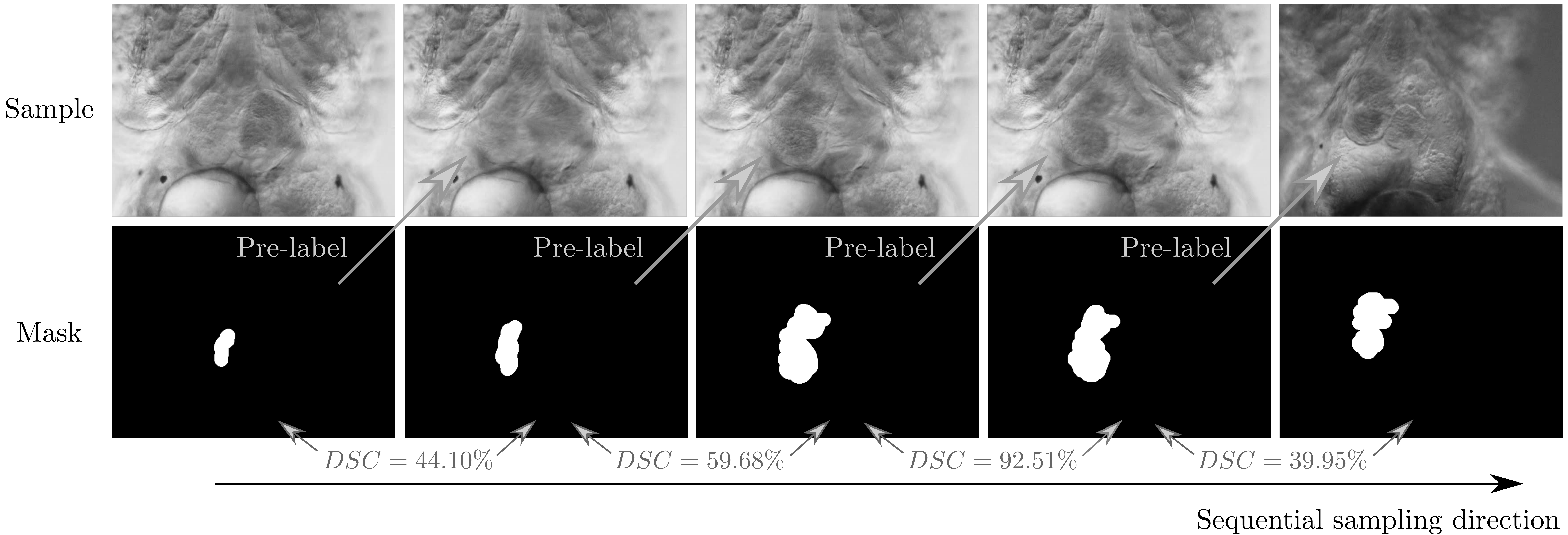}
	}
	\caption{Comparison of sample, corresponding mask, and $DSC$ between neighboring frames to illustrate temporal pre-labeling (sequential sampling form left to right) on the medaka dataset.}
	\label{fig:pre-labeling-temporal}%
\end{figure}
Firstly, the potential of the proposed previous label usage is analyzed at the medaka dataset since it is composed of video sequences like presented in Section~\ref{sec:datasets}. Figure~\ref{fig:pre-labeling-temporal} illustrates a sequence of the sequential sampling and used pre-labels. In addition to the visual impression, $DSC$~\eqref{eq:dsc} is printed to compare neighboring label masks. It can be shown that the first three pre-labels are beneficial since there is a direct relation between frames. Consequently, $DSC$ is larger than 40\% in each of those frames. Especially, frames 3 and 4 are very similar, which can be demonstrated by a $DSC = 92.51 \%$. However, the last frame illustrates a remaining problem in the method if sequences change. Hereby, the displayed pre-label is not helpful in order to do image annotation of the last sample. 
\begin{figure}
	\centering
	\resizebox{.8\textwidth}{!}{%
		\includegraphics[]{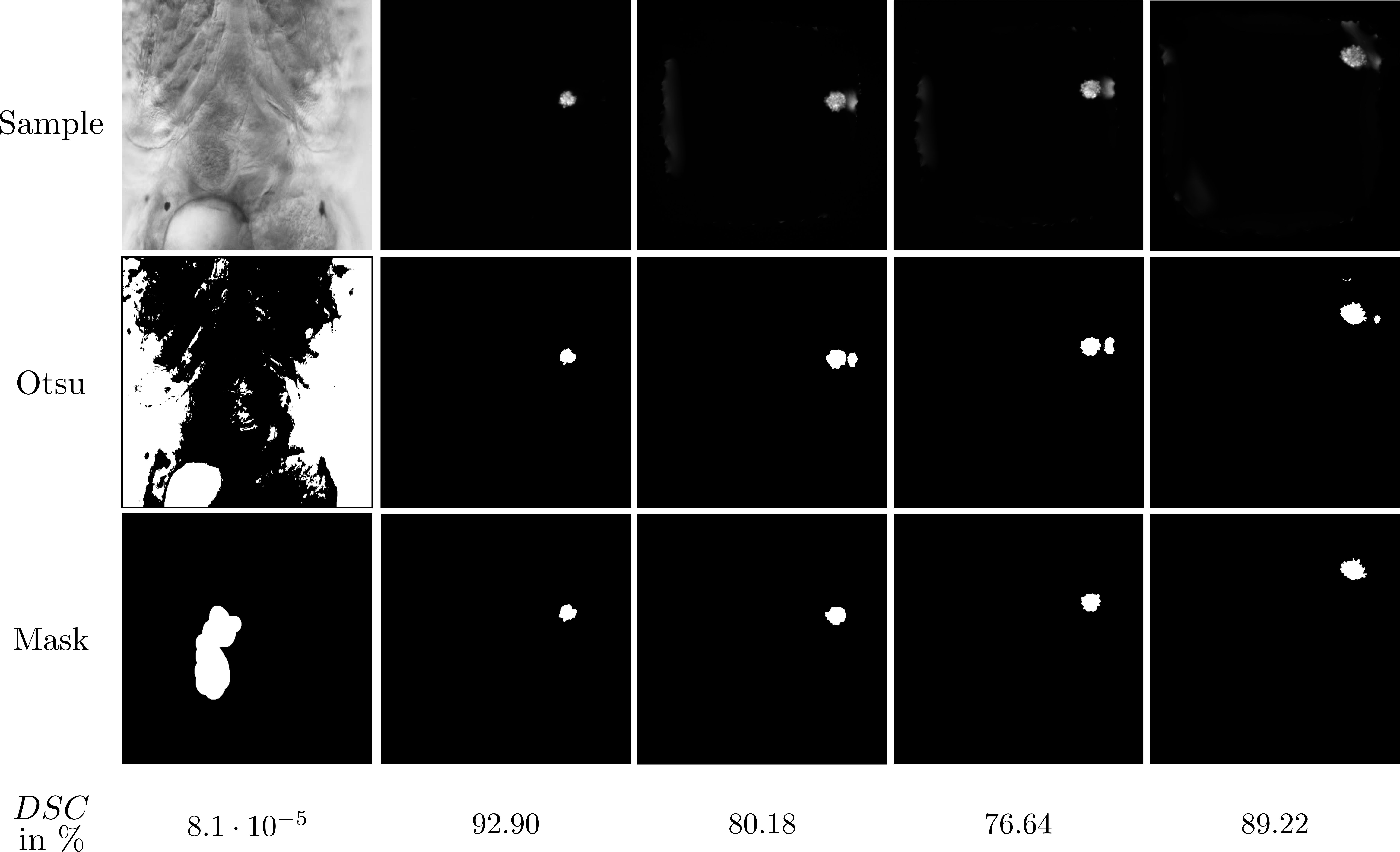}
	}
	\caption{Illustration of visual differences between Otsu pre-labeling and ground truth mask as well as $DSC$ to quantify the similarity of masks on medaka (first column) and DMA spheroid samples (remaining columns).}
	\label{fig:pre-labeling-otsu}%
\end{figure}
Figure~\ref{fig:pre-labeling-otsu} presents pre-labeling using Otsu thresholding~\cite{Otsu1979ThresholdSelectionMethod}. In order to execute Otsu on RGB medaka images, an upstream transformation to a gray-level image space is done at first. However, the algorithm is not suitable as a pre-labeling strategy for medaka images, which, in addition to visual inspection, a $DSC$ tending to zero demonstrates, too. Thus, in this case, pre-labeling would impede annotation instead of simplifying it. Nevertheless, Otsu performs very well on DMA samples shown by $DSC\geq 76\%$. Hence, it provides helpful initial guesses w.r.t. DMA data. Having a closer inspection and comparing it with the ground truth masks, it can be recognized, that there are still small wrong mask segments. However, deleting the wrong mask segment, in this case, is much more efficient than starting image annotation from scratch. The main reason is that curved boundaries of the spheroid are already correctly predicted for the most part.

\begin{figure}
	\centering
	\resizebox{0.75\textwidth}{!}{%
		\input{./figs-src/unet-pre-labeling.tex}
	}
	\caption{Illustration of DNN pre-labeling performance: comparison different amounts of training data ($\mid\mathcal{D}^l_{\text{train}}\mid$) w.r.t. visual impression and $DSC$ between ground truth mask and pre-labels respectively DNN predictions.}%
	\label{fig:pre-labeling-DNN}%
\end{figure}
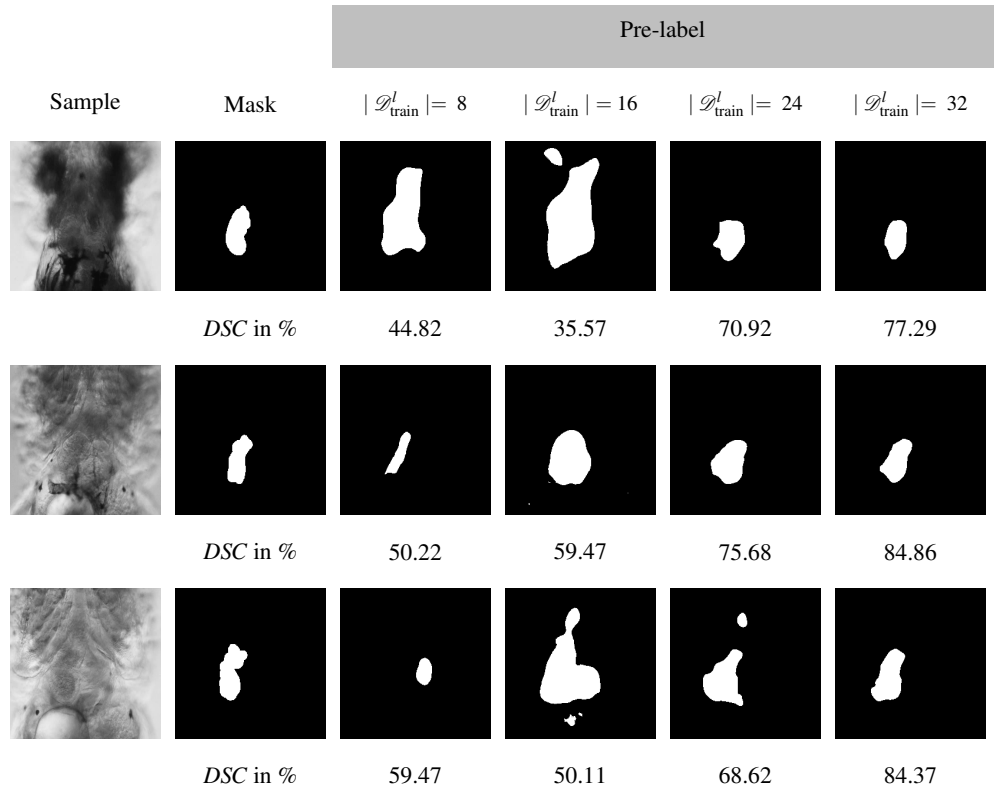
Since there is no obvious heuristic for medaka dataset, we investigate how DNN U-Net trained on a small labeled dataset can be used as pre-labeling. Results for different amounts of training data $\mid\mathcal{D}^l_{\text{train}}\mid$ following random sampling presented in Section~\ref{sec:method-selector} can be found in Figure~\ref{fig:pre-labeling-DNN}. We compare pre-labels and ground truth masks of samples $\mathbf{x} \notin  \mathcal{D}^l_{\text{train}}$ not represented in the training dataset by visual impression and $DSC$~\eqref{eq:dsc} in parallel. Our experiments show, that by using only $\mid\mathcal{D}^l_{\text{train}}\mid=~32$ labels, a DNN can serve as a meaningful and generic pre-label strategy on medaka dataset. Furthermore, we offer in our tool the opportunity to export a training job that can directly be sent to data scientists to avoid the requirements of a graphics processing unit on the labeling device. Hence, the annotator only needs to select DNN weights provided by a data scientist. The inference time on the introduced hardware (Intel i3-1110G4) of $t_{\text{inference}} = 0.75$ s is a feasible processing amount during labeling.  
\paragraph{Human Annotator User Experience}
We have presented our implemented software prototype nested in a touchscreen device to several users and have requested feedback concerning labeling comfort. The overall feedback of users has been positive. Most of the users named a comfort enhancement during image annotation using a touchscreen. However, very experienced users w.r.t. mouse labeling remark that for them touchscreen labeling is not superior to using a mouse as an input device since they are used to it. Thus, especially for an average user labeling via touchscreen may facilitate access to the procedure. Concluding results, several possibilities of user input maintain the maximum level of adaption to the needs of users. 
\paragraph{Post-Assistance}
\begin{figure}
	\centering
	\begin{subfigure}[c]{0.445\textwidth}
		\resizebox{\textwidth}{!}{%
			\centering
			\includegraphics{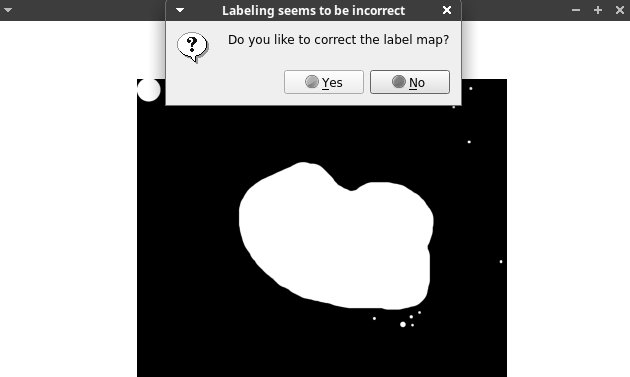}
		}
		\subcaption{Label inspection.}
		\label{fig:label-inspector}
	\end{subfigure}	
	\hfill
	\begin{subfigure}[l]{0.43\textwidth}
		\resizebox{\textwidth}{!}{%
			\centering
			\includegraphics{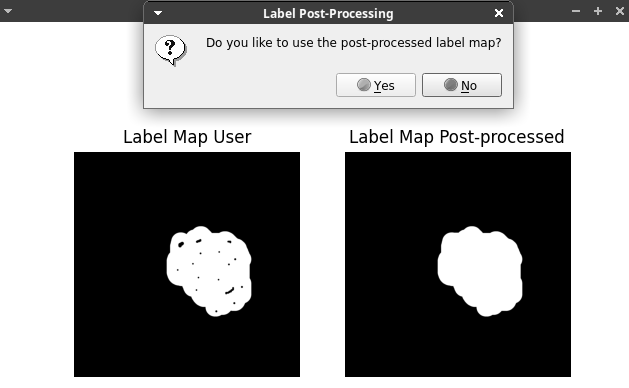}
		}
		\subcaption{Label post-processing.}
		\label{fig:post-processing}
	\end{subfigure}	
	\caption{Examples of post-assistance: (a) an inspector warns the user since there is more than one segment labeled, (b) a post-processing can be performed to fill holes.}
	\label{fig:post-assist}%
\end{figure}
Figure~\ref{fig:label-inspector} illustrates a label inspection evaluating deviations of connected segments to the desired segment number as a quality metric $\gamma_i$ introduced in Equation \eqref{eq:gamma}. Large deviations lead to the presented warning prompt and give users the possibility to relabel images. Consequently, using the feedback mechanism can help to increase attention w.r.t. noisy labels directly during annotation. Post-processing links reoccurring errors with an opportunity to straightforwardly solve them. Figure~\ref{fig:post-processing} presents post-processing in form of closing intending to avoid holes in segment masks. Similar to label inspection, the annotator can adopt the post-processing suggestion or reject it avoiding unwanted changes. Therefore, post-processing enables a way of handling common error sources using algorithms like morphological operations or custom functions depending on the underlying problem.

The key results can be summarized the following: A selector can help to reduce the amount of labeled data needed to achieve accurate DNN results. Pre-processing and pre-labeling can facilitate annotation and decrease the effort needed for labeling an image. Human annotators gain more flexibility by providing different types of input devices. Label inspection and post-processing build awareness of label quality and ways to deal with it. 
\section{Conclusion}
\label{sec:conclusion}
Dealing with Deep Learning (DL), labeling plays an important role. We motivated that assisting annotators during labeling is desired (reducing labeling effort and increasing label quality). Methods to tackle these issues are various, but a summary and combination of those in a general concept is lack. We contribute a summary of properties and challenges in datasets w.r.t. annotation. Besides, we propose a generic workflow combing and extending various ideas of labeling enhancement. Especially, an evolved concept of label inspection and post-processing implemented directly within the annotation process is presented as a novel way to increase label quality. Our contribution is intended to serve as a template, which can be used by the community for practical DL projects where a labeled dataset is required. To make this concept applicable, we present a software prototype implementation as an initial starting point that can be customized. Several functionalities are demonstrated using the prototype processing two biomedical image segmentation datasets. The prototype enables further research on enhancing image annotation and investigations of new underlying methods like more generic feedback approaches or active learning in the proposed pipeline modules. For instance, the initial required amount of labeled data or further quantification of enhancement using an assisted labeling approach may be part of further research.
\section*{Acknowledgement}
This work was funded in the KIT Future Fields project "Screening Platform for Personalized Oncology (SPPO)" and was performed on the computational resource bwUniCluster 2.0 funded by the Ministry of Science, Research and the Arts Baden-Württemberg and the Universities of the State of Baden-Württemberg, Germany, within the framework program bwHPC.

%% file: figs-src/state-of-the-art-labeling.tex
\usetikzlibrary{shapes.geometric}
\tikzset{database/.style={cylinder,aspect=0.5,draw,rotate=90,path picture={
			\draw (path picture bounding box.160) to[out=180,in=180] (path picture bounding
			box.20);
			\draw (path picture bounding box.200) to[out=180,in=180] (path picture bounding
			box.340);
}}}
\begin{tikzpicture}[
every node/.append style={font={\sffamily},inner sep=2pt},
caption/.append style={font={\sffamily\bfseries}, inner sep=0pt},
comm/.style={-latex,semithick,inner sep=2pt},
msg/.style={midway,sloped,above,font={\sffamily\small},inner sep=2pt},
inner sep=0pt
]
\matrix[
row sep=1.5ex,
column sep=6mm,
nodes={ 
	line width=1pt,
	anchor=center, 
	text centered,
	rounded corners,
	minimum width=0.7cm, minimum height=10mm}] 
{%
	\node[cylinder, rotate=90, draw, rounded corners=0, minimum size=0.8cm](data-unlabeled){}; & \node[draw](annotator) {\includegraphics[width=2.7cm]{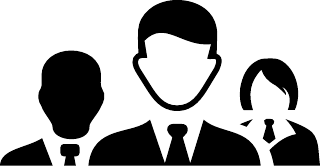}}; & \node[cylinder, rotate=90, draw, rounded corners=0, minimum size=0.8cm, fill=lightgray](data-labeled){}; \\
	\node {Unlabeled dataset $\mathcal{D}^{\text{u}}$};  &  \node {Human annotator};  & \node () {Labeled dataset $\mathcal{D}^{\text{l}}$};\\
};
\draw[comm] (data-unlabeled)--(annotator) node[msg, xshift=-0.5cm] (test) {$\mathbf{x}$};
\draw[comm] (annotator)--(data-labeled) node[msg] {$\mathbf{y}$};

\node [right of=data-unlabeled, node distance=1.4cm, inner sep=0, outer sep = 0]  (start) {};
\draw[comm] (start) |- ++(1,1.4) -| (data-labeled);

\end{tikzpicture}

%% file: figs-src/workflow.tex
\begin{tikzpicture}[
every node/.append style={font={\sffamily},inner sep=2pt},
caption/.append style={font={\sffamily\bfseries}, inner sep=0pt},
comm/.style={-latex,semithick,inner sep=2pt},
msg/.style={midway,sloped,above,font={\sffamily\small},inner sep=2pt},
inner sep=0pt
]
\matrix[
row sep=1.5ex,
column sep=6mm,
nodes={ 
	line width=1pt,
	anchor=center, 
	text centered,
	rounded corners,
	minimum width=0.7cm, minimum height=10mm}] 
{%
	\node{};  &  \node {\includegraphics[width=1.2cm]{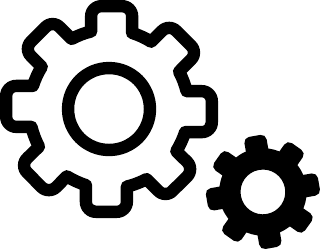}}; & \node {\includegraphics[width=1.2cm]{./figs/gear.pdf}}; & \node {\includegraphics[width=1.2cm]{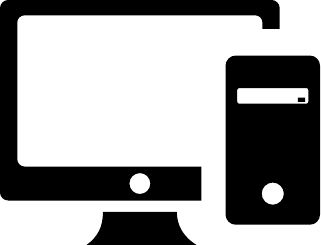}}; & \node {\includegraphics[width=1.2cm]{./figs/gear.pdf}};& \node{}; \\
	\node{};  &  \node(selector-methods) {Methods}; & \node(pre-methods) {Methods}; & \node(devices)  {Devices}; & \node(post-methods) {Methods}; & \node{}; \\
	\node[cylinder, rotate=90, draw, rounded corners=0, minimum size=0.8cm](data-unlabeled){};  &  \node[draw](selector) {\includegraphics[width=1.2cm]{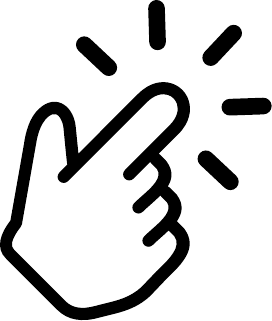}}; &  \node[draw](pre) {\includegraphics[width=1.85cm]{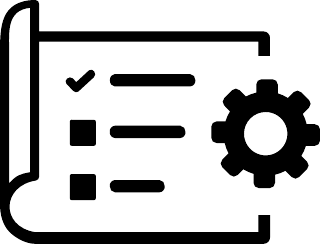}}; & \node[draw](annotator) {\includegraphics[width=2.7cm]{./figs/annotator.pdf}}; & \node[draw](post) {\includegraphics[width=1.85cm]{./figs/processing.pdf}}; & \node[cylinder, fill=lightgray, rotate=90, draw, rounded corners=0, minimum size=0.8cm](data-labeled){}; \\
	\node {Unlabeled dataset $\mathcal{D}^{\text{u}}$};  &  \node {Selector}; &  \node {Pre-Assistance}; &  \node {Human annotator}; & \node {Post-Assistance}; & \node () {Labeled dataset $\mathcal{D}^{\text{l}}$};\\
};

\draw[comm] (data-unlabeled)--(selector) node[msg] {};
\draw[comm] (selector)--(pre) node[msg, xshift=-.1cm] {$\mathbf{x}^{*}$};
\node [right of=selector, node distance=1.075cm, inner sep=0, outer sep = 0]  (start) {};
\draw[comm] (start) |- ++(2,4) -| (data-labeled);
\draw[comm] (pre)--(selector) node[msg] {};

\draw[comm] (pre)--(annotator) node[msg] {};
\draw[comm] (annotator)--(post) node[msg] {};
\draw[comm] (post)--(annotator) node[msg] {};
\draw[comm] (post)--(data-labeled) node[msg] {$\mathbf{y}$};

\draw[comm] (selector-methods)--(selector) node[msg] {};
\draw[comm] (pre-methods)--(pre) node[msg] {};
\draw[comm] (post-methods)--(post) node[msg] {};
\draw[comm] (devices)--(annotator) node[msg] {};
\end{tikzpicture}

%% file: figs-src/medaka-dataset.tex
\begin{tikzpicture}[
caption/.append style={font={\sffamily\bfseries}, inner sep=0pt},
comm/.style={-latex,semithick,inner sep=2pt},
msg/.style={midway,sloped,above,font={\sffamily\small},inner sep=2pt},
inner sep=0pt
]
\matrix[
row sep=1mm,
column sep=.1mm,
nodes={ 
	line width=1pt,
	anchor=center, 
	text centered,
	rounded corners,
	minimum width=2.1cm, minimum height=8mm}] 
{%
	\node{ Sample};  &  \node{\includegraphics[width=2cm, height=2cm]{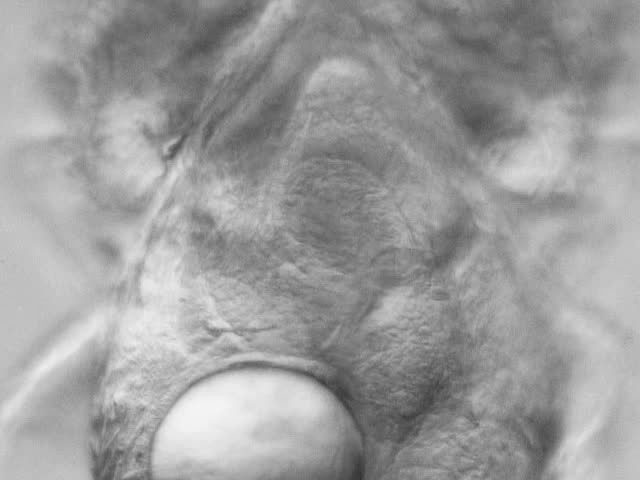}};  & \node{\includegraphics[width=2cm, height=2cm]{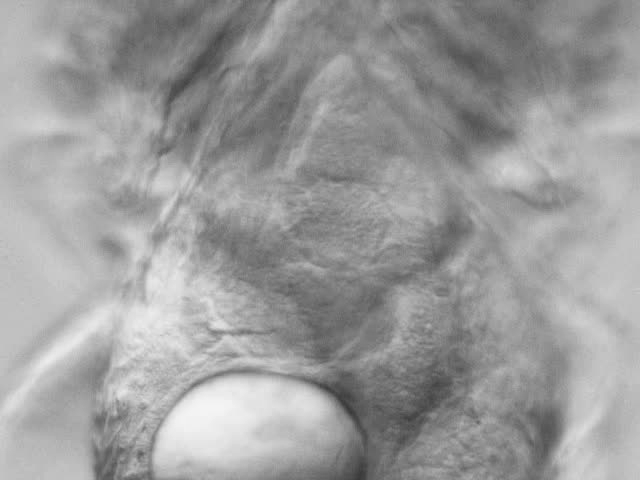}}; &  \node{\includegraphics[width=2cm, height=2cm]{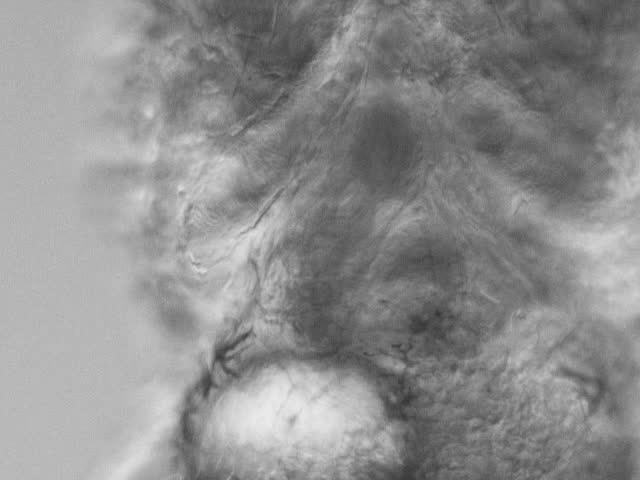}}; &  \node{\includegraphics[width=2cm, height=2cm]{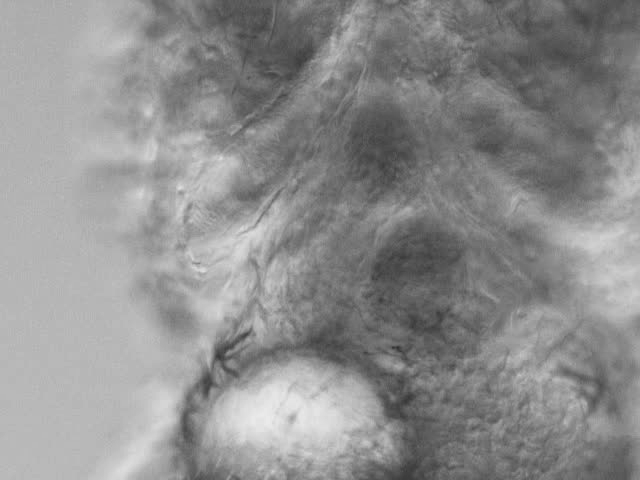}}; \\
	\node { Mask}; & \node{\includegraphics[width=2cm, height=2cm]{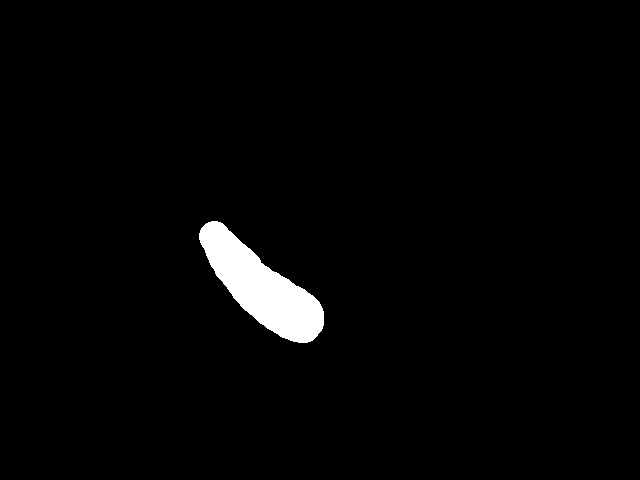}};  & \node{\includegraphics[width=2cm, height=2cm]{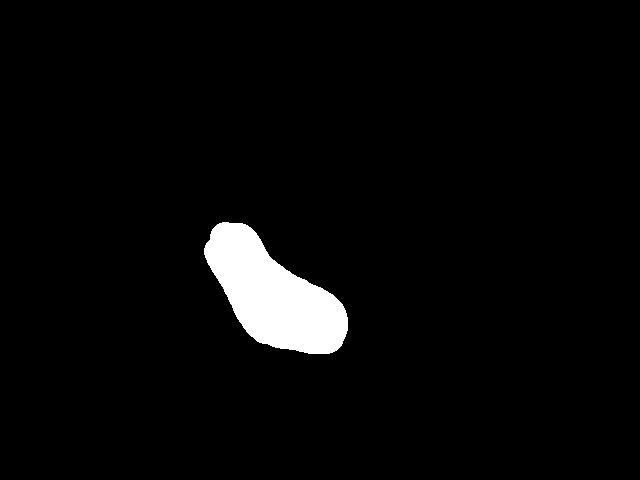}}; &  \node{\includegraphics[width=2cm, height=2cm]{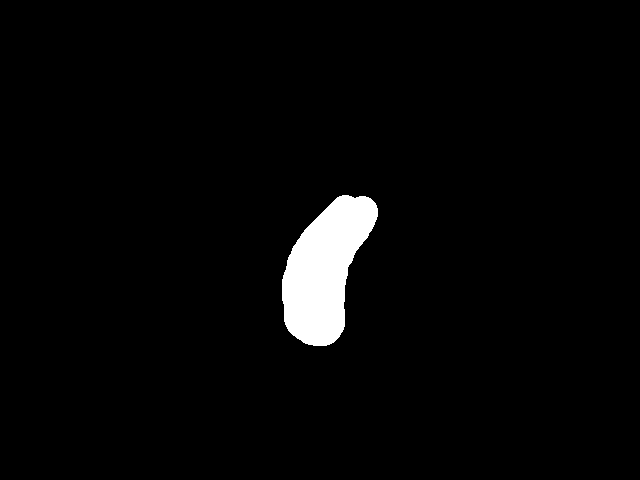}}; &  \node{\includegraphics[width=2cm, height=2cm]{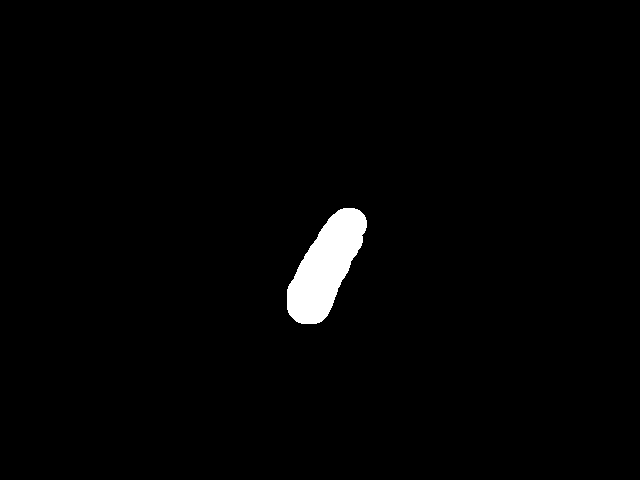}};\\
};
\end{tikzpicture}

%% file: figs-src/dma-dataset.tex
\begin{tikzpicture}[
caption/.append style={font={\sffamily\bfseries}, inner sep=0pt},
comm/.style={-latex,semithick,inner sep=2pt},
msg/.style={midway,sloped,above,font={\sffamily\small},inner sep=2pt},
inner sep=0pt
]
\matrix[
row sep=1mm,
column sep=.1mm,
nodes={ 
	line width=1pt,
	anchor=center, 
	text centered,
	rounded corners,
	minimum width=2.1cm, minimum height=8mm}] 
{%
	\node{Sample};  &  \node{\includegraphics[width=2cm, height=2cm]{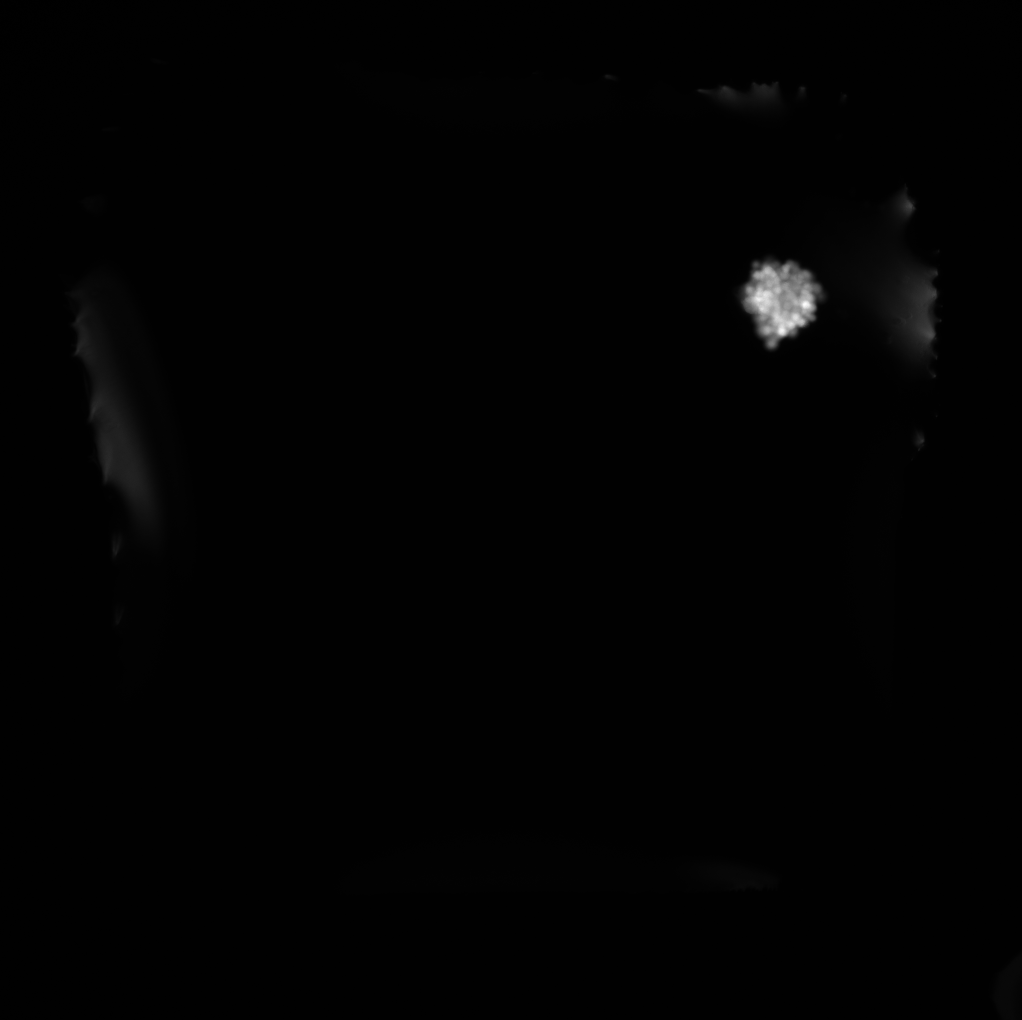}};  & \node{\includegraphics[width=2cm, height=2cm]{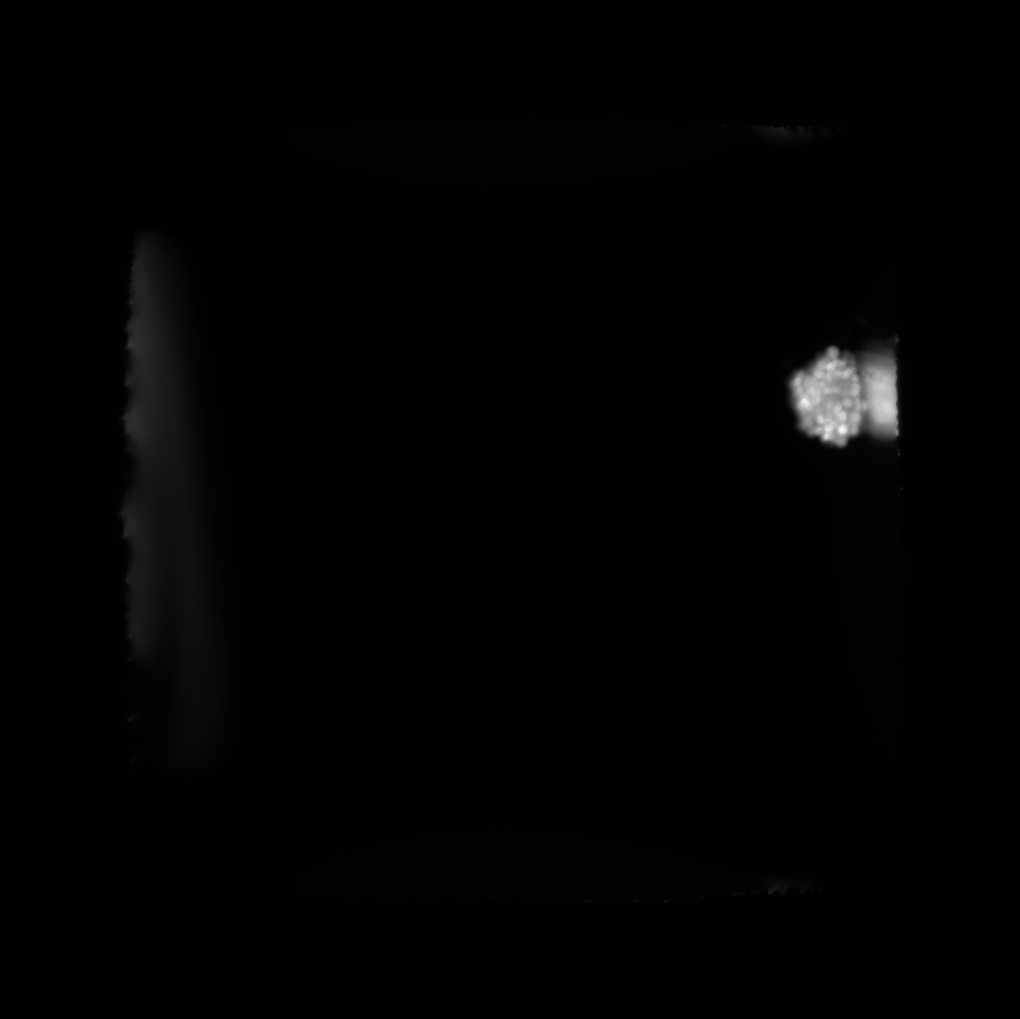}}; &  \node{\includegraphics[width=2cm, height=2cm]{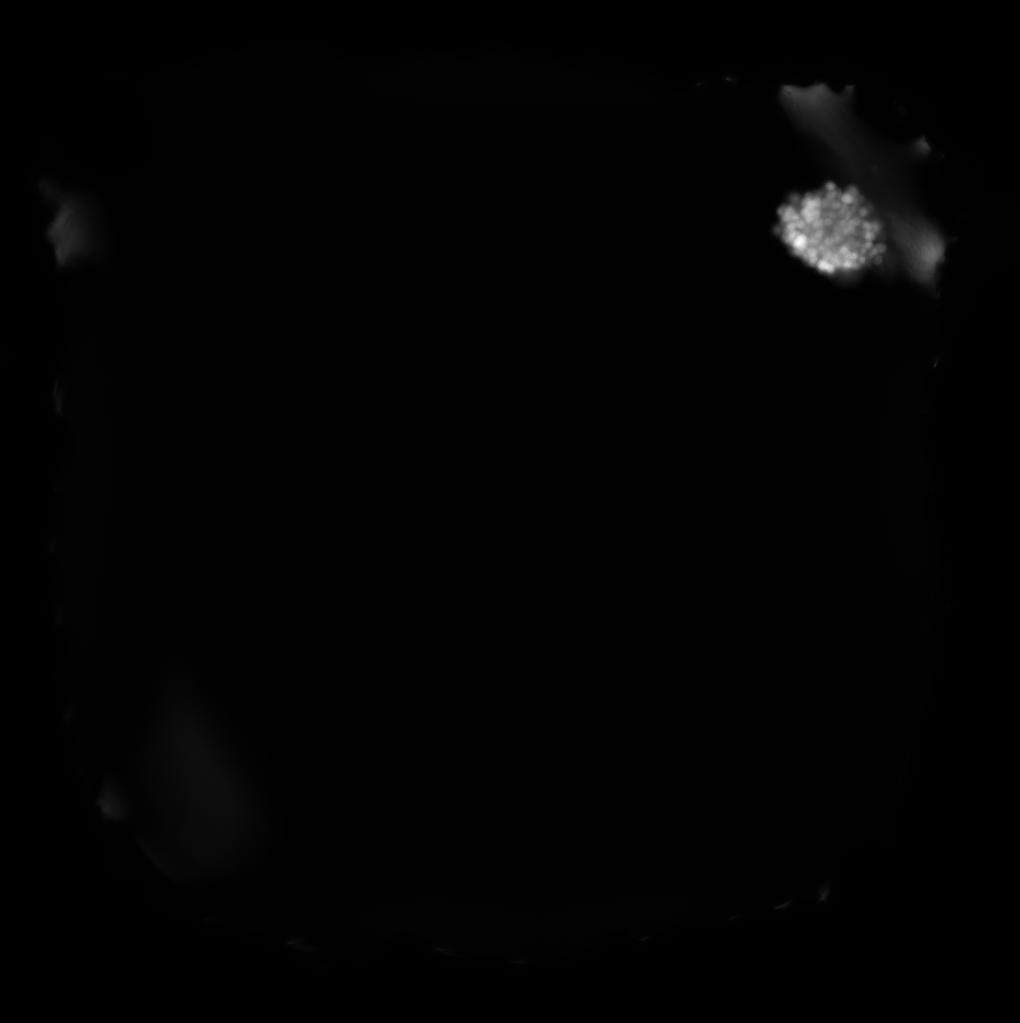}}; &  \node{\includegraphics[width=2cm, height=2cm]{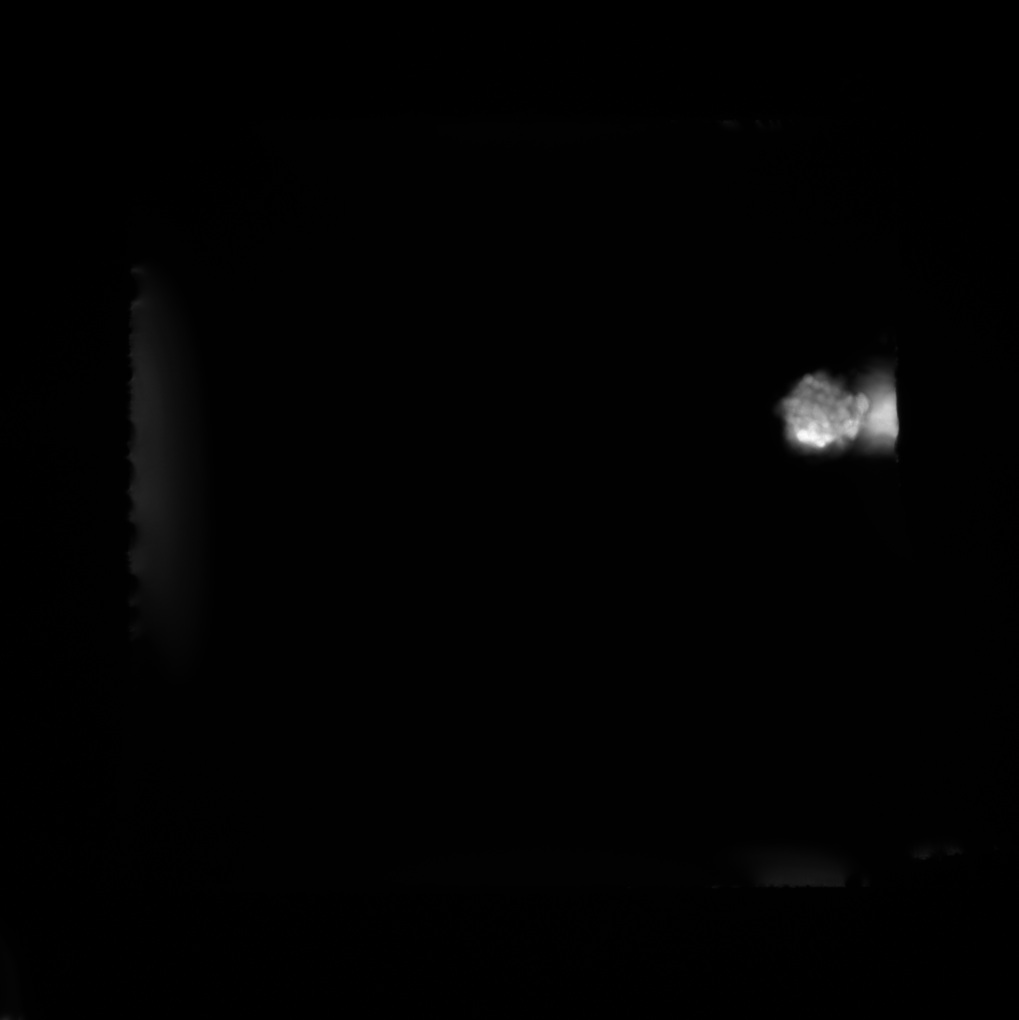}}; \\
	\node {Mask}; & \node{\includegraphics[width=2cm, height=2cm]{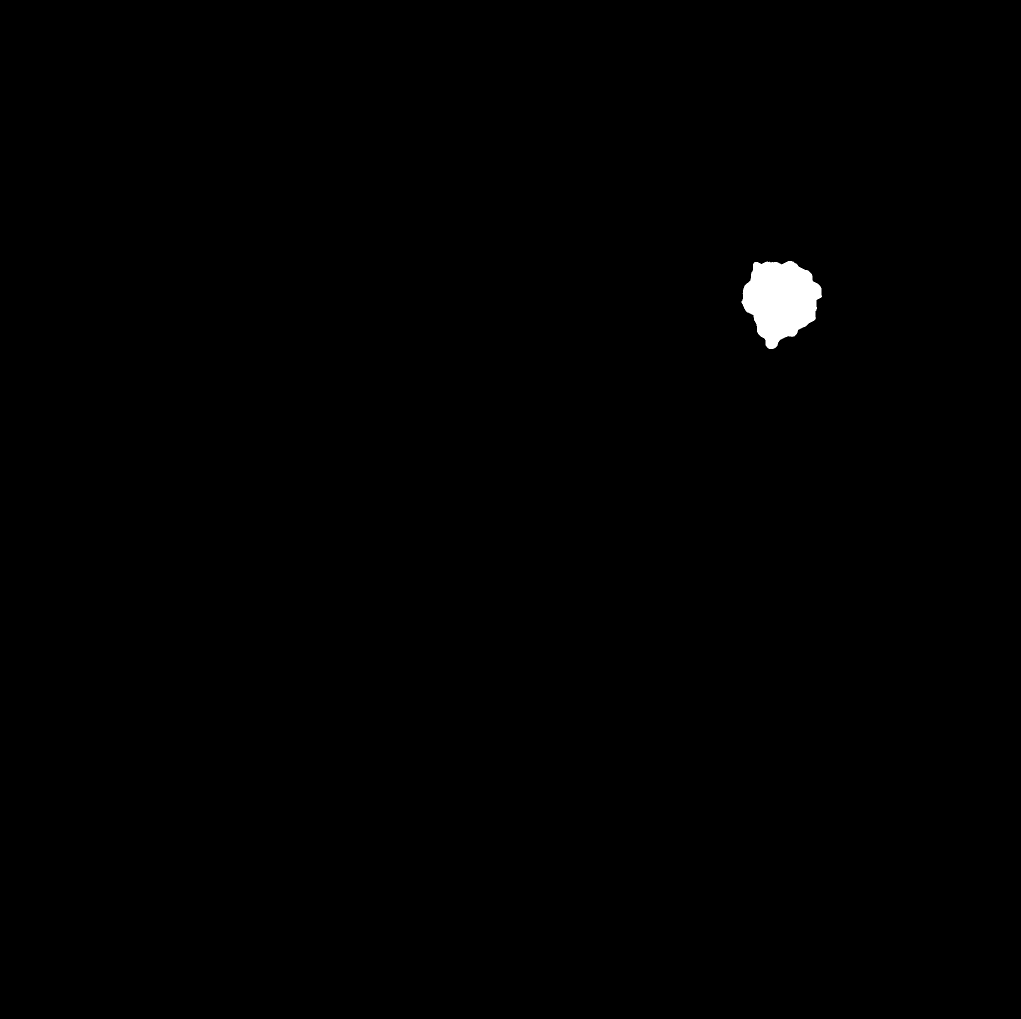}};  & \node{\includegraphics[width=2cm, height=2cm]{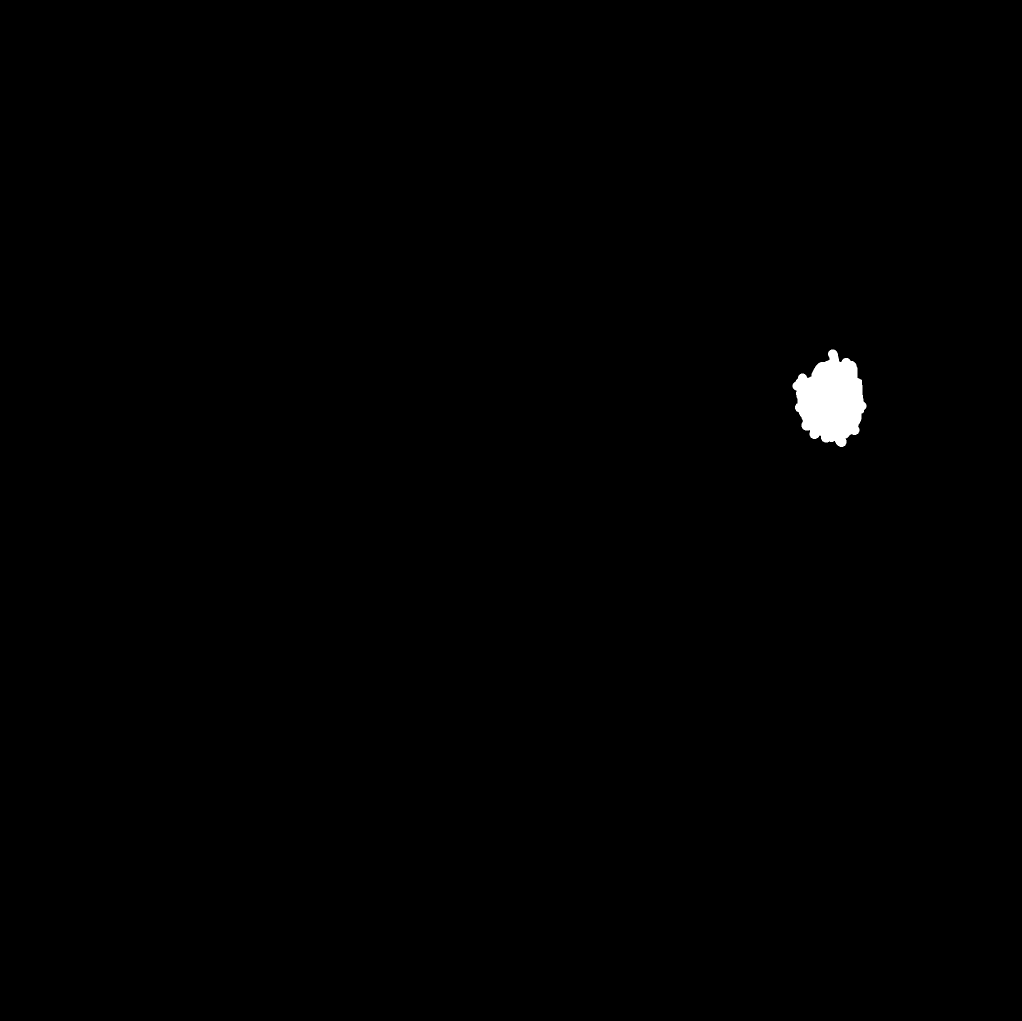}}; &  \node{\includegraphics[width=2cm, height=2cm]{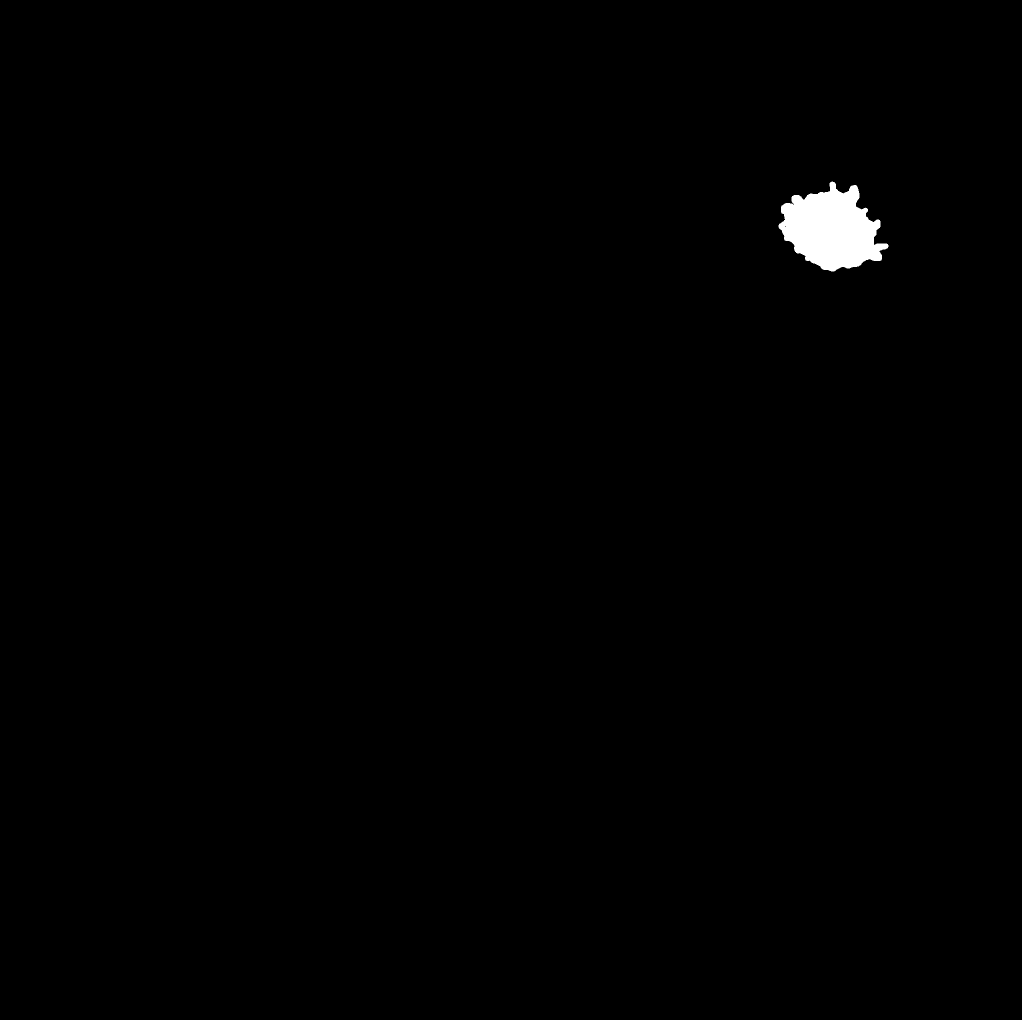}}; &  \node{\includegraphics[width=2cm, height=2cm]{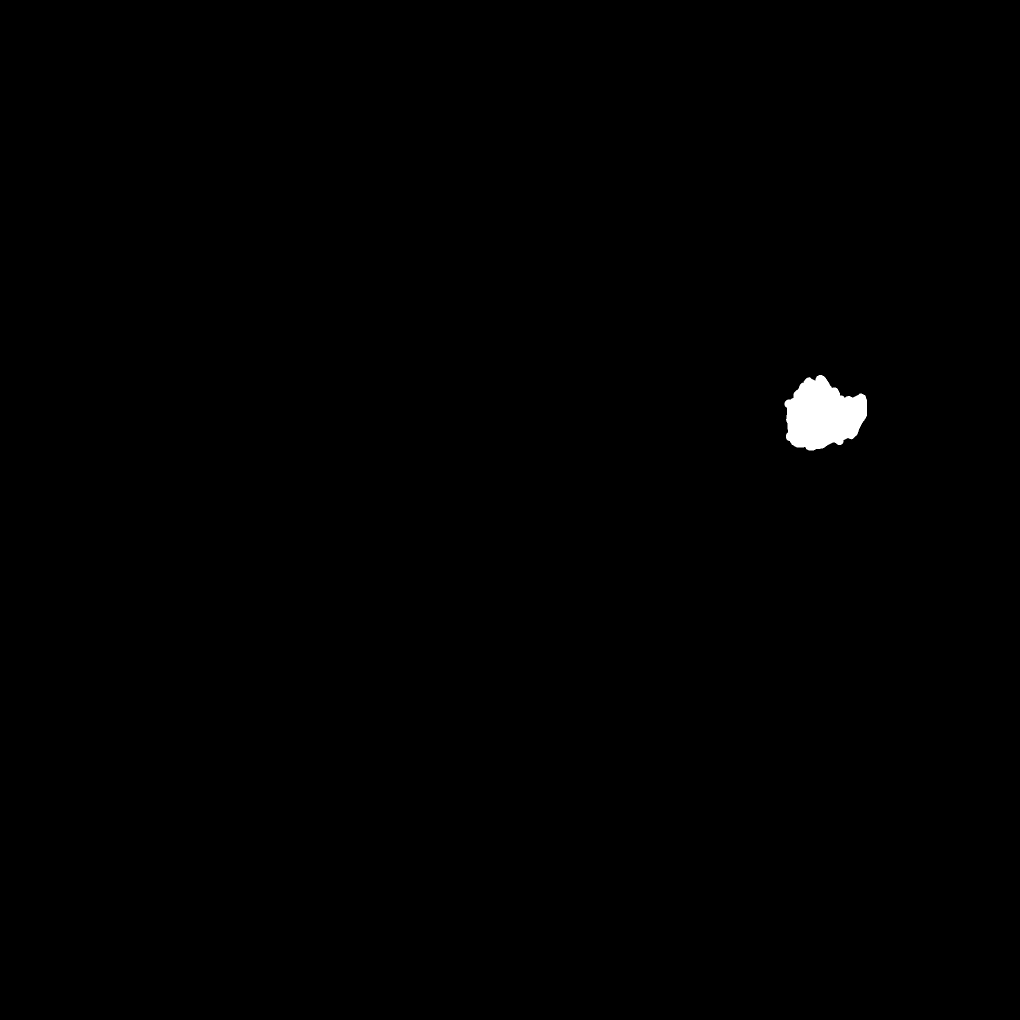}};\\
};
\end{tikzpicture}

%% file: figs-src/unet-pre-labeling.tex
\begin{tikzpicture}[
caption/.append style={font={\sffamily\bfseries}, inner sep=0pt},
comm/.style={-latex,semithick,inner sep=2pt},
msg/.style={midway,sloped,above,font={\sffamily\small},inner sep=2pt},
inner sep=0pt
]
\matrix(table)[
row sep=1mm,
column sep=.1mm,
nodes={ 
	line width=1pt,
	anchor=center, 
	text centered,
	rounded corners,
	minimum width=2.2cm, minimum height=8mm}] 
{%
	\node{};  &  \node {}; & \node(beg) {}; &  \node {}; &  \node {};&  \node(end) {};\\
	\node{Sample};  &  \node {Mask}; & \node {\footnotesize $\mid\mathcal{D}^l_{\text{train}}\mid=~8$}; &  \node {\footnotesize$\mid\mathcal{D}^l_{\text{train}}\mid~=16$}; &  \node {\footnotesize$\mid\mathcal{D}^l_{\text{train}}\mid=~24$};&  \node {\footnotesize$\mid\mathcal{D}^l_{\text{train}}\mid=~32$};\\
	\node{\includegraphics[width=2cm, height=2cm]{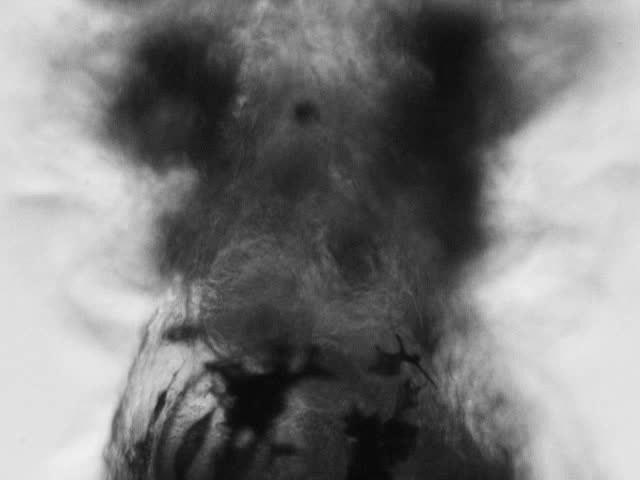}};  &  \node {\includegraphics[width=2cm, height=2cm]{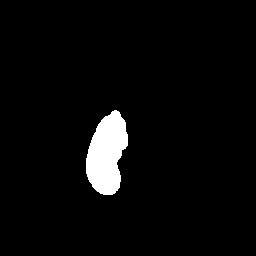}}; & \node {\includegraphics[width=2cm, height=2cm]{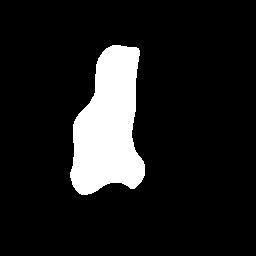}}; & \node {\includegraphics[width=2cm, height=2cm]{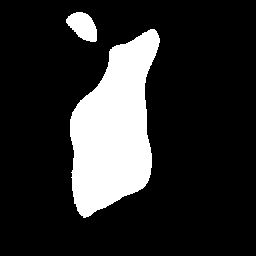}}; & \node {\includegraphics[width=2cm, height=2cm]{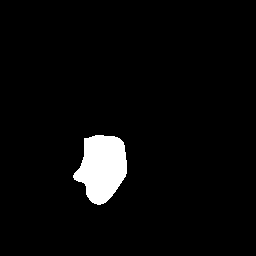}};& \node {\includegraphics[width=2cm, height=2cm]{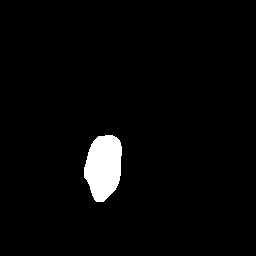}}; \\
	\node{};  &  \node{$DSC$ in \%}; & \node {44.82}; & \node{35.57}; & \node{70.92};& \node {77.29}; \\
	\node{\includegraphics[width=2cm, height=2cm]{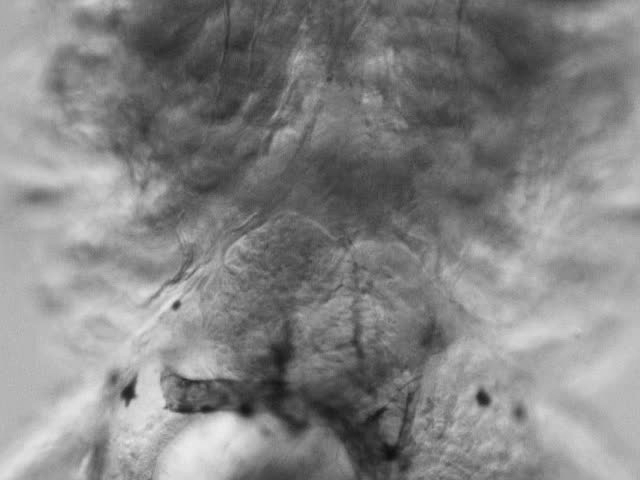}};  &  \node {\includegraphics[width=2cm, height=2cm]{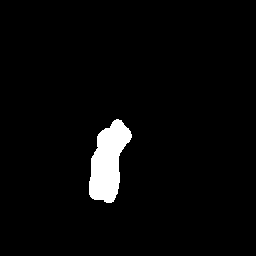}}; & \node {\includegraphics[width=2cm, height=2cm]{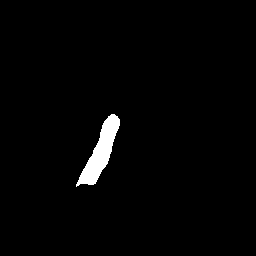}}; & \node {\includegraphics[width=2cm, height=2cm]{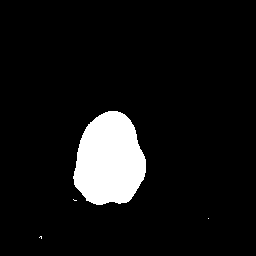}}; & \node {\includegraphics[width=2cm, height=2cm]{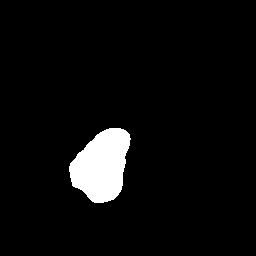}};& \node {\includegraphics[width=2cm, height=2cm]{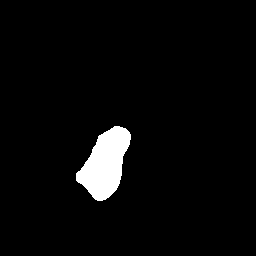}}; \\
	\node{};  &  \node{$DSC$ in \%}; & \node {50.22}; & \node{59.47}; & \node{75.68};& \node {84.86}; \\
	\node{\includegraphics[width=2cm, height=2cm]{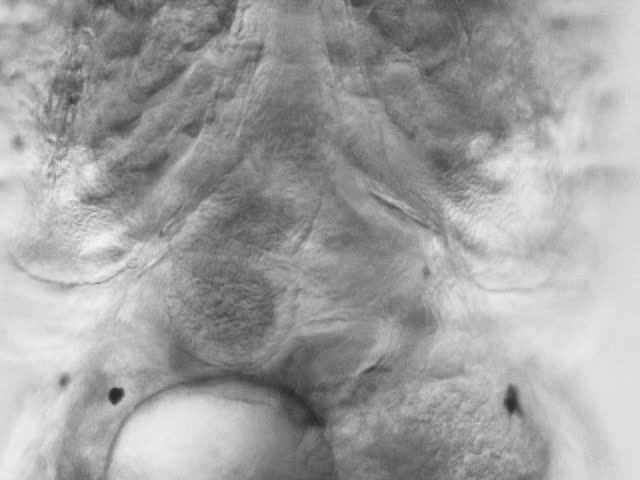}};  &  \node {\includegraphics[width=2cm, height=2cm]{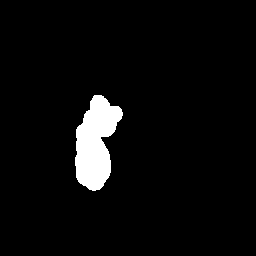}}; & \node {\includegraphics[width=2cm, height=2cm]{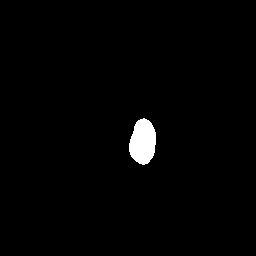}}; & \node {\includegraphics[width=2cm, height=2cm]{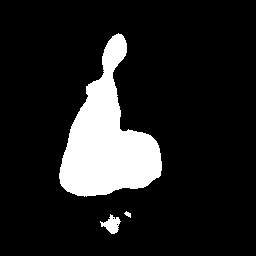}}; & \node {\includegraphics[width=2cm, height=2cm]{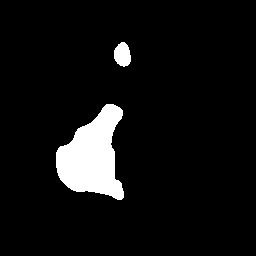}};& \node {\includegraphics[width=2cm, height=2cm]{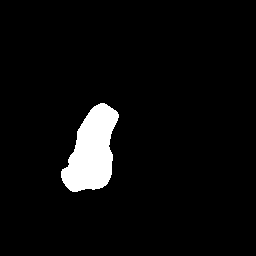}}; \\
	\node{};  &  \node{$DSC$ in \%}; & \node {59.47}; & \node{50.11}; & \node{68.62};& \node {84.37}; \\
};
    \node[fit=(beg)(end), fill=lightgray, minimum height=1mm]{Pre-label};
\end{tikzpicture}